\def\arborpart{full}
\documentclass[10pt,twocolumn,letterpaper]{article}

\usepackage[pagenumbers]{cvpr}      %

\usepackage[utf8]{inputenc}
\usepackage[T1]{fontenc}
\usepackage{amsfonts}
\usepackage{nicefrac}
\usepackage{microtype}         %
\usepackage{capt-of}
\usepackage{array}
\usepackage{tabularx}
\usepackage{makecell}
\usepackage{colortbl}
\usepackage{tikz}
\def\arborpartmain{main}\def\arborpartsupp{supp}
\ifx\arborpart\arborpartmain
  \usepackage{xr-hyper}
\fi
\ifx\arborpart\arborpartsupp
  \usepackage{xr-hyper}
\fi

\definecolor{cvprblue}{rgb}{0.21,0.49,0.74}
\usepackage[pagebackref,breaklinks,colorlinks,allcolors=cvprblue]{hyperref}
\hypersetup{pdfauthor={Jan-Niklas Dihlmann, Andreas Engelhardt, Simon Donne, Hendrik P. A. Lensch, Mark Boss},pdftitle={Arbor: Explicit Geometric Conditioning for Controllable 3D Asset Generation}}

\def\confName{CVPR}
\def\confYear{2027}

\newcommand{\inputlocal}[1]{}

\title{Arbor: Explicit Geometric Conditioning\\for Controllable 3D Asset Generation}
\author{%
  Jan-Niklas Dihlmann\textsuperscript{1,2}\quad
  Andreas Engelhardt\textsuperscript{2}\quad
  Simon Donn\'e\textsuperscript{2}\quad
  Hendrik P.\,A.\ Lensch\textsuperscript{1}\quad
  Mark Boss\textsuperscript{2}\\
  {\normalsize\textsuperscript{1}University of T\"ubingen\qquad\textsuperscript{2}Stability AI\qquad
  \url{https://arbor.jdihlmann.com}}\vspace{-0.6em}
}

\definecolor{hull}{HTML}{7EE165}
\definecolor{avoidance}{HTML}{FE0000}
\definecolor{touch}{HTML}{FFBE00}
\definecolor{mismatch}{HTML}{6FB2F7}

\definecolor{revclaritycol}{HTML}{1F5FBF}   %
\definecolor{revdeltacol}{HTML}{8E24AA}     %
\definecolor{revfigurecol}{HTML}{00897B}    %
\definecolor{revevalcol}{HTML}{E65100}      %
\definecolor{revdetailcol}{HTML}{6D4C41}    %
\definecolor{revscopecol}{HTML}{827717}     %
\definecolor{citeissuecol}{HTML}{D50000}    %

\ifdefined\localdraft
  \newcommand{\aen}[1]{{\color{blue}AE:~#1}}
  \newcommand{\mb}[1]{{\color{teal}MB:~#1}}
  \newcommand{\todoopen}[1]{{\color{red}TODO:~#1}}
  \newcommand{\reviewerrisk}[1]{{\color{orange}Reviewer risk:~#1}}
  \newcommand{\evidencegap}[1]{{\color{red}Evidence gap:~#1}}
  \newcommand{\revnote}[4]{{\color{#1}\sffamily\footnotesize\textbf{[#2 $\vert$ #3]}~#4\textbf{]}}}
  \newcommand{\revclarity}[2]{}
  \newcommand{\revdelta}[2]{}
  \newcommand{\revfigure}[2]{}
  \newcommand{\reveval}[2]{}
  \newcommand{\revdetail}[2]{}
  \newcommand{\revscope}[2]{}
  \newcommand{\citeissue}[2]{{\color{citeissuecol}\sffamily\footnotesize\textbf{[CITATION $\vert$ #1]}~#2\textbf{]}}}
\else
  \newcommand{\aen}[1]{}
  \newcommand{\mb}[1]{}
  \newcommand{\todoopen}[1]{}
  \newcommand{\reviewerrisk}[1]{}
  \newcommand{\evidencegap}[1]{}
  \newcommand{\revnote}[4]{}
  \newcommand{\revclarity}[2]{}
  \newcommand{\revdelta}[2]{}
  \newcommand{\revfigure}[2]{}
  \newcommand{\reveval}[2]{}
  \newcommand{\revdetail}[2]{}
  \newcommand{\revscope}[2]{}
  \newcommand{\citeissue}[2]{}
\fi

\newcommand{\inlinesection}[1]{\vspace{0.05cm}\noindent\textbf{#1}}

\newcommand{\step}[1]{\tikz[baseline=(s.base)]{\node[circle,fill=black!55,text=white,inner sep=0.5pt,minimum size=0.95em,font=\sffamily\bfseries\scriptsize](s){#1};}}
\ifx\arborpart\arborpartmain
  \newcommand{\extref}[1]{\ref*{#1}}
\else
  \newcommand{\extref}[1]{\ref{#1}}
\fi
\ifdefined\localdraft
  \newcommand{\todonum}[1]{{\setlength{\fboxsep}{1pt}\colorbox{yellow}{\sffamily\footnotesize ?\,#1}}}
\else
  \newcommand{\todonum}[1]{\textbf{??}}
\fi

\begin{document}

\ifx\arborpart\arborpartsupp
  \appendix
\providecommand{\changenote}[1]{\leavevmode{\color{magenta}\sffamily\footnotesize[#1]}}
\providecommand{\figslot}[2]{\begingroup\setlength{\fboxsep}{0pt}\fcolorbox{black!12}{black!4}{\parbox[c][#1][c]{\dimexpr\linewidth-2\fboxrule\relax}{\centering\color{black!45}\sffamily\footnotesize #2}}\endgroup}
\newcommand{\suppsummary}[1]{\par\vspace{0.1em}\noindent\colorbox{cvprblue!8}{\parbox{\dimexpr\columnwidth-2\fboxsep\relax}{\small\textbf{Summary.} #1}}\par\vspace{0.5em}}
\newif\ifsuppshowslots\suppshowslotstrue
\ifsuppshowslots
\newcommand{\suppplaceholder}[5][\textwidth]{%
\begin{figure*}[!tbp]
    \centering
    \figslot{0.2\textwidth}{\textbf{Figure in preparation}\\[2pt]#3}
    \caption{\textbf{#3.}}
    \label{#2}
\end{figure*}}
\fi
\setcounter{table}{0}\setcounter{figure}{0}
\renewcommand{\thetable}{S\arabic{table}}\renewcommand{\thefigure}{S\arabic{figure}}
\makeatletter\@ifundefined{theHtable}{}{\renewcommand{\theHtable}{S\arabic{table}}\renewcommand{\theHfigure}{S\arabic{figure}}}\makeatother

\newcommand{\vidpaper}{\texttt{explain\_video.mp4}}
\newcommand{\vidtrailer}{\texttt{trailer\_video.mp4}}
\newcommand{\vidrow}[2]{\hspace*{0.8em}#1 \hfill {\footnotesize\texttt{#2}}}

\newcommand{\supphero}[1]{%
    \centering
    \includegraphics[width=\textwidth]{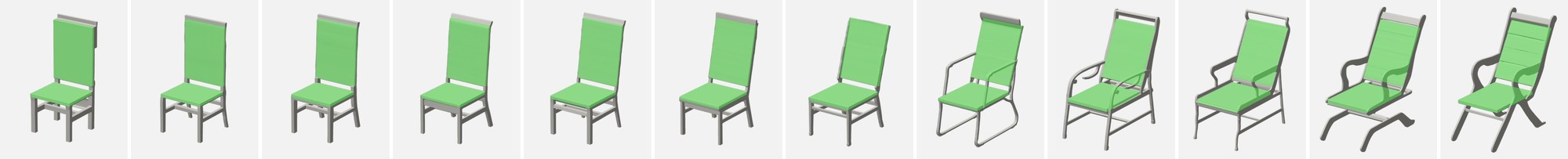}
    #1{\textbf{A constraint sweep.}
    Arbor generates twelve chairs for the prompt ``Chair''. From left to right the hull (green) of the seat and the back tilts further backward.}
    \label{fig:supp:videos}}
  \raggedbottom
  \twocolumn[{%
    \centering
    {\Large \textbf{\thetitle}\\[0.5em] Supplementary Material\par}
    \vspace{1.2em}
    \supphero{\captionof{figure}}
    \vspace{1.2em}
  }]
\refstepcounter{section}\label{app:supplement}\label{app:guide}

\subsection*{Overview}
In this supplementary material we show further results of Arbor and provide the details behind the experiments of the main paper.
Several videos accompany the paper on the project page, listed below.
We highly recommend watching the explain video first, since it presents the whole paper in an easy to follow form.

\vspace{0.3em}
\noindent\colorbox{black!5}{\parbox{\dimexpr\columnwidth-2\fboxsep\relax}{\small
\textbf{Videos on the project page}\\[1pt]{\footnotesize\url{https://arbor.jdihlmann.com}}\\[3pt]
Explain video, narrated\\[1pt]
Trailer\\[1pt]
Chairs of Fig.~\ref{fig:introduction:overview}\\[1pt]
Turntables of Fig.~\ref{fig:comparison}\\[1pt]
Image input, Fig.~\ref{fig:recon_comparison}\\[1pt]
Decoder and probe, Fig.~\ref{fig:decoder_detail}\\[1pt]
Constraint sweeps of Fig.~\ref{fig:constraint_sweep} and Fig.~\ref{fig:supp:additional_sweeps}}}
\vspace{1.6em}

\noindent\textbf{Contents}\par\vspace{0.4em}
{\small
\newcommand{\tocrow}[3]{\textbf{\ref{#1}} & \textbf{#2} & \pageref{#1} \\[-1pt] & {\footnotesize #3} & \\[4pt]}
\noindent\begin{tabular*}{\columnwidth}{@{\extracolsep{\fill}}@{}l@{\hspace{0.8em}}lr@{}}
\tocrow{app:data_constraints}{Data and Benchmark}{Training data, constraint sampling, benchmark splits}
\tocrow{app:extended_results}{Additional Qualitative Results}{Further objects, baselines, prompts, constraint sweeps}
\tocrow{app:more_numbers}{Additional Quantitative Results}{Confidence intervals, score variants, surface detail}
\tocrow{app:variants}{Design Choices}{Sampling time guidance, encoder, training loss}
\tocrow{app:router_k}{Router Capacity and Cost}{Token cap of the router, runtime, memory}
\tocrow{app:implementation_details}{Implementation Details}{Guidance, model, training, decoder adapters}
\tocrow{app:metrics}{Evaluation Details}{Metric definitions, baselines, user study}
\tocrow{app:limitations}{Limitations}{Resolution, soft conditioning, constraint types}
\tocrow{app:qa}{Questions and Answers}{Short answers with pointers to the evidence}
\end{tabular*}}

\subsection{Data and Benchmark}\label{app:data_constraints}

\suppsummary{We train Arbor on about $50$k objects. We generate the constraints on the fly from each training object, so we need no annotated constraints. For the evaluation we use objects from Toys4K, which Arbor never sees in training. The automatic split has $128$ sampled constraint sets, and the manual split has $32$ sets that we made by hand in Blender.}

\paragraph{Training data.}
We train Arbor on objects from ABO~\citep{abo2022}, HSSD~\citep{hssd2024}, and the Sketchfab subset of Objaverse-XL~\citep{objaversexl2023} (Tab.~\ref{tab:dataset_summary}).
This is about a tenth of the data that TRELLIS~\citep{trellis2024} was trained on.
A smaller set is enough here, since the generator stays frozen and only the attachment has to be trained.
All training settings are listed in Sec.~\ref{app:implementation_details} (Tab.~\ref{tab:supp:settings}).

\begin{table}[t]
    \centering
    \small
    \renewcommand{\arraystretch}{1.08}
    \setlength{\tabcolsep}{5pt}
    \begin{tabular*}{\columnwidth}{@{\extracolsep{\fill}}llr@{}}
        \toprule
        Dataset & Role & Objects \\
        \midrule
        ABO~\citep{abo2022} & training & 4{,}484 \\
        HSSD~\citep{hssd2024} & training & 6{,}660 \\
        Objaverse-XL~\citep{objaversexl2023} & training & 39{,}846 \\
        \midrule
        Toys4K~\citep{toys4k2021} & benchmark & $128 + 32$ \\
        \bottomrule
    \end{tabular*}
    \caption{\textbf{Data sources.}
    For training we count the objects with a part segmentation.
    For the benchmark we count the objects of the automatic and the manual split.}
    \label{tab:dataset_summary}
\end{table}

\begin{figure*}[t]
    \centering
    \includegraphics[width=\textwidth]{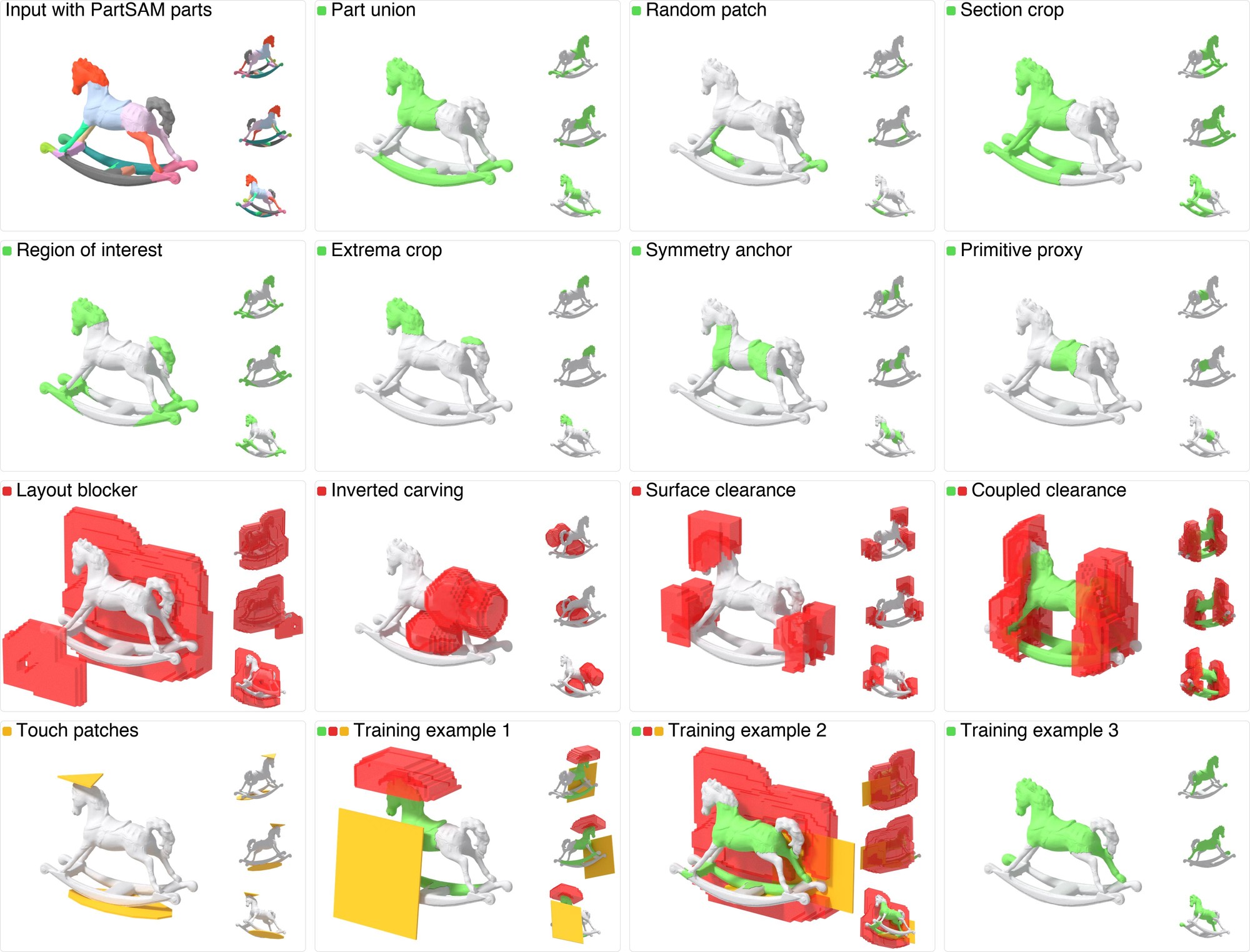}
    \caption{\textbf{The twelve constraint samplers.}
    We show every sampler on one training object, with \textcolor[rgb]{0.34,0.85,0.32}{hull} samplers in green, \textcolor[rgb]{0.9,0.18,0.18}{avoidance} samplers in red and the \textcolor[rgb]{0.95,0.7,0.1}{touch} sampler in yellow.
    The first cell shows its PartSAM~\citep{partsam2026} parts, and the last three cells are full training examples.}
    \label{fig:supplement:constraint_families}
\end{figure*}

\begin{figure*}[!tbp]
    \centering
    \includegraphics[width=\textwidth]{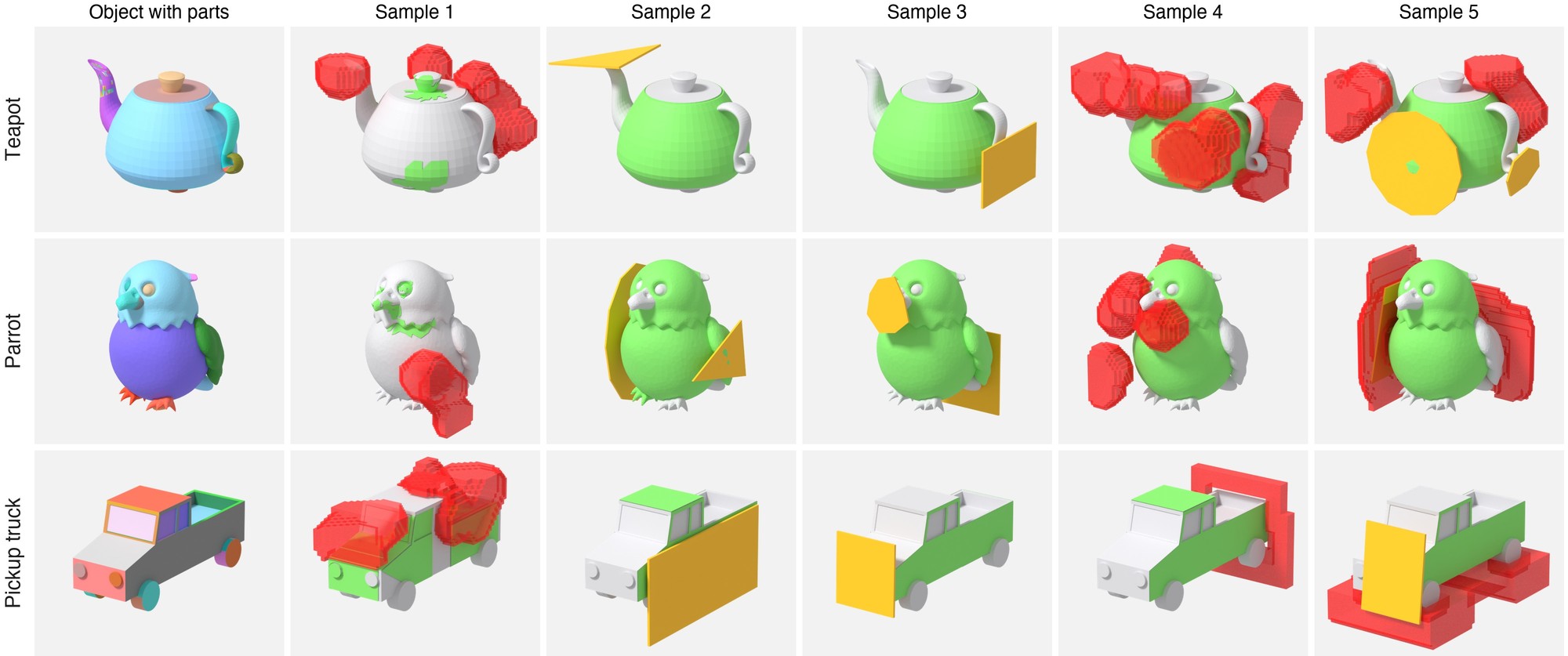}
    \caption{\textbf{Training constraints on different objects.}
    Each row shows one object with its parts and five constraint sets that the sampler draws for it during training.}
    \label{fig:supplement:constraint_families_objects}
\end{figure*}

\paragraph{Constraint sampler.}
There is no dataset that pairs meshes with hull, avoidance and touch regions, so we create this data ourselves.
We treat the constraints as an augmentation that happens on the fly.
Our data loader loads a mesh and then applies a set of rules that cut hull, avoidance and touch regions from it.
Since every constraint comes from the training object itself, the object is always a valid answer to its constraint.
The model also sees a new constraint every time an object is drawn.
We use twelve such rules, which we call samplers, in three families (Fig.~\ref{fig:supplement:constraint_families}), and we list them with their weights in Tab.~\ref{tab:supp:samplers}.
The hull samplers select geometry that the object should keep.
In most cases they keep whole parts from an offline PartSAM~\citep{partsam2026} segmentation, such as the head or the runners of a rocking horse.
The other hull samplers grow a patch on the surface, crop a half or a slab of the object, or keep the surface inside a primitive.
The avoidance samplers place regions that have to stay empty.
These regions are blockers around the object, blobs in its free space, or clearance in front of a surface.
The touch sampler places patches at the sides of the object, which the object has to reach from one side.
Every training example contains a hull, and we add avoidance and touch regions in half of the examples each.
In Fig.~\ref{fig:supplement:constraint_families_objects} we show five such examples for each of three further objects.
The same object gives a different constraint with every draw, from a single part up to a set that combines all three types.

\paragraph{Object frame.}
We define all constraints in the normalized object frame of TRELLIS, which is a cube with side length one around the object.
This means that a constraint describes positions relative to the object and not in a scene.
An object that is meant for a scene is first generated in this frame and then placed in the scene.

\par\medskip\noindent\begin{minipage}{\columnwidth}
    \centering
    \footnotesize
    \renewcommand{\arraystretch}{1.05}
    \setlength{\tabcolsep}{4pt}
    \begin{tabular*}{\columnwidth}{@{\extracolsep{\fill}}llc@{}}
        \toprule
        Type & Sampler & Weight \\
        \midrule
        Hull, in every example & Part union & $0.6$ \\
         & Each of the six others & $0.067$ \\
        \midrule
        Avoidance, in half of the examples & Layout blocker & $0.35$ \\
         & Inverted carving & $0.35$ \\
         & Surface clearance & $0.2$ \\
         & Coupled clearance & $0.1$ \\
        \midrule
        Touch, in half of the examples & Touch patches & $1$ \\
        \bottomrule
    \end{tabular*}
    \captionof{table}{\textbf{Constraint samplers and their weights.}
    One sampler is drawn per active type with the given weight.}
    \label{tab:supp:samplers}
\end{minipage}\par\medskip

\begin{figure*}[!tbp]
    \centering
    \includegraphics[width=\textwidth]{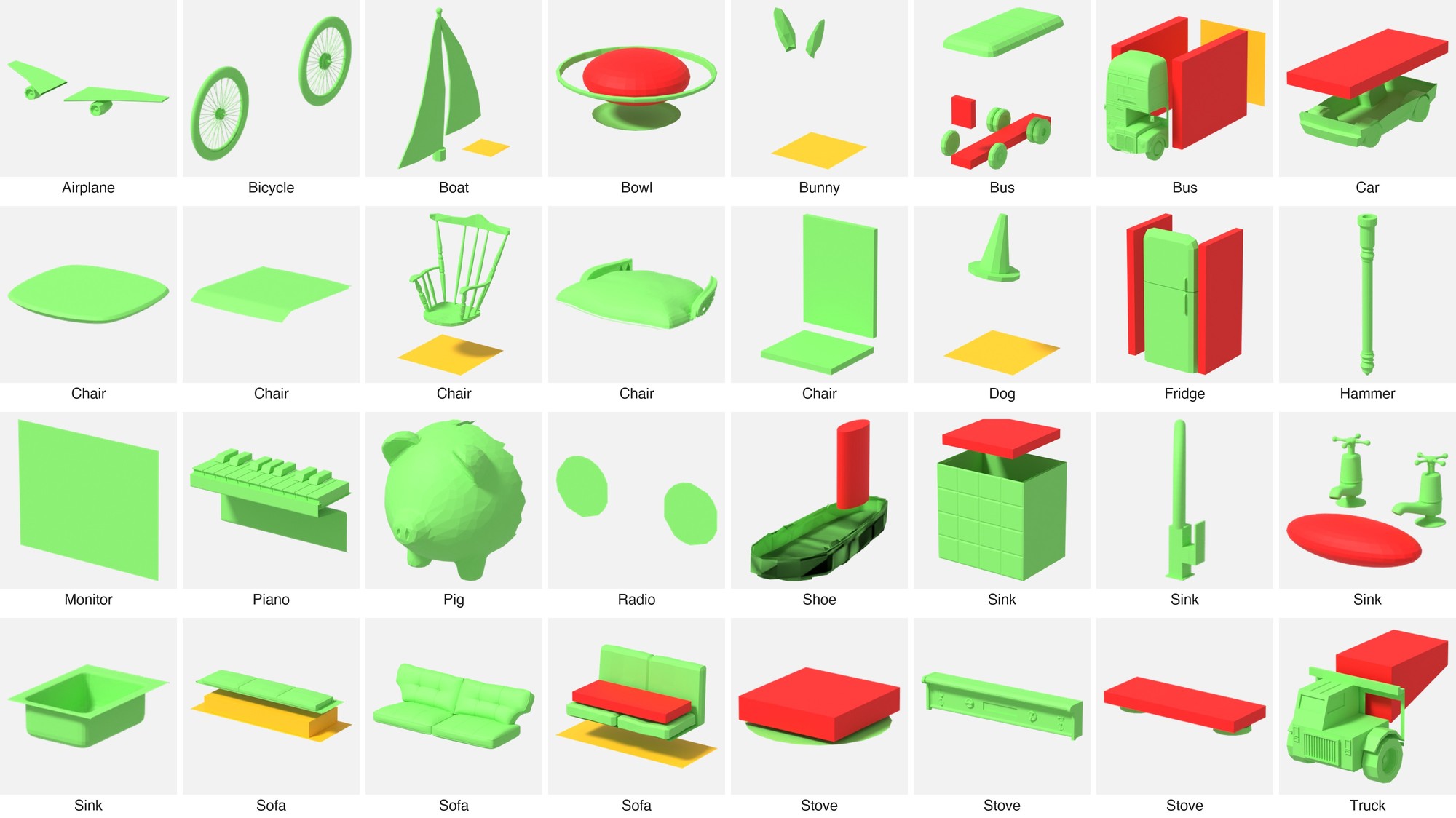}
    \caption{\textbf{The $32$ constraint sets of the manual split.}
    We modeled each set by hand with the intent of an artist in mind.
    The name below a cell is the category of the object.
    Colors as in Fig.~\ref{fig:supplement:constraint_families}.}
    \label{fig:supp:manual_sets}
\end{figure*}

\renewcommand{\dblfloatpagefraction}{0.85}
\begin{figure*}[!tbp]
    \centering
    \includegraphics[width=\textwidth]{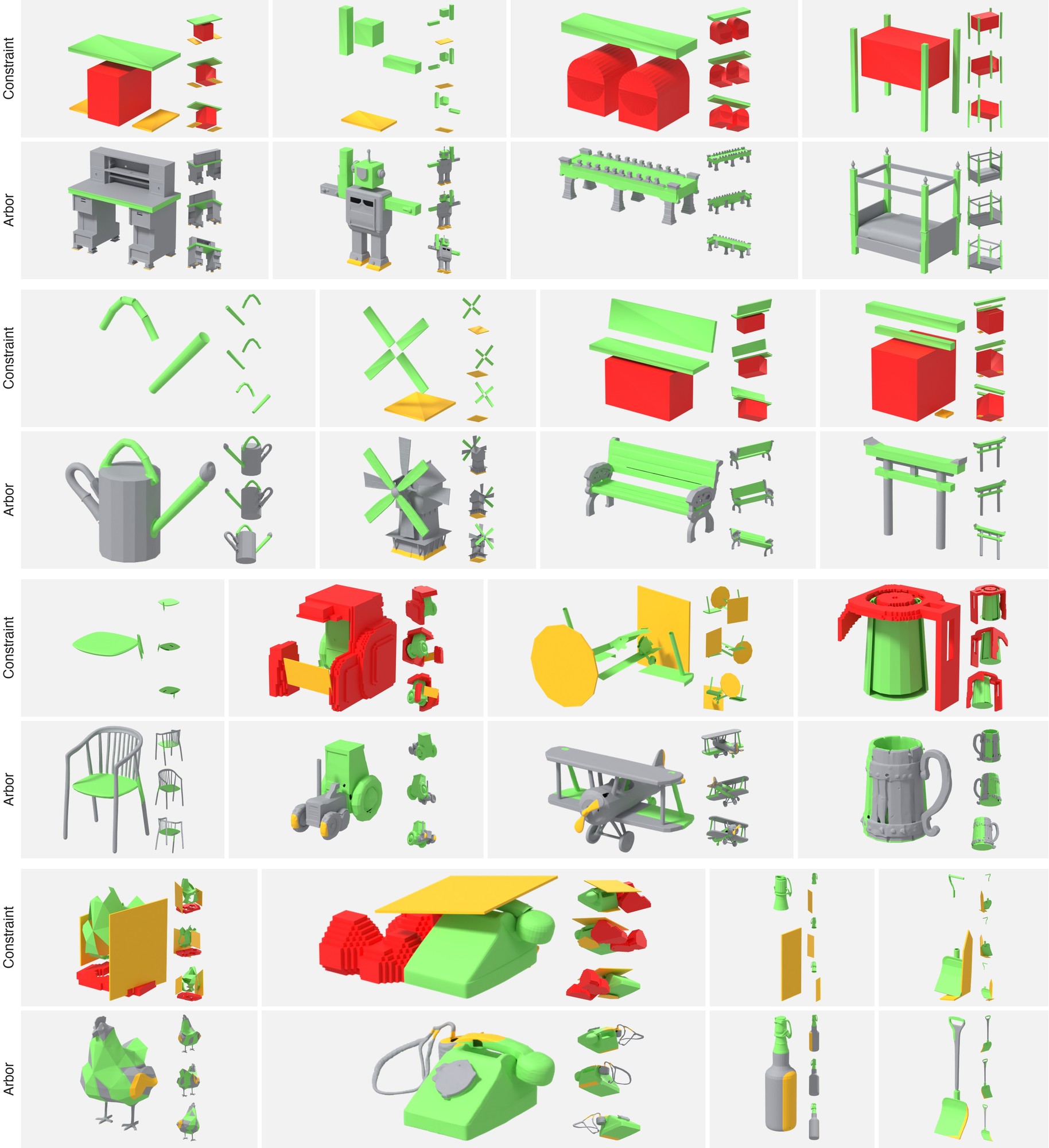}
    \caption{\textbf{Additional results of Arbor.}
    Layout and colors as in Fig.~\ref{fig:comparison}.
    In the top two blocks the constraints are built from boxes, cylinders and ellipsoids without a reference object, and the bottom two blocks use sets of the automatic split.}
    \label{fig:supp:gallery2}
\end{figure*}

\begin{figure*}[!tbp]
    \centering
    \includegraphics[width=\textwidth]{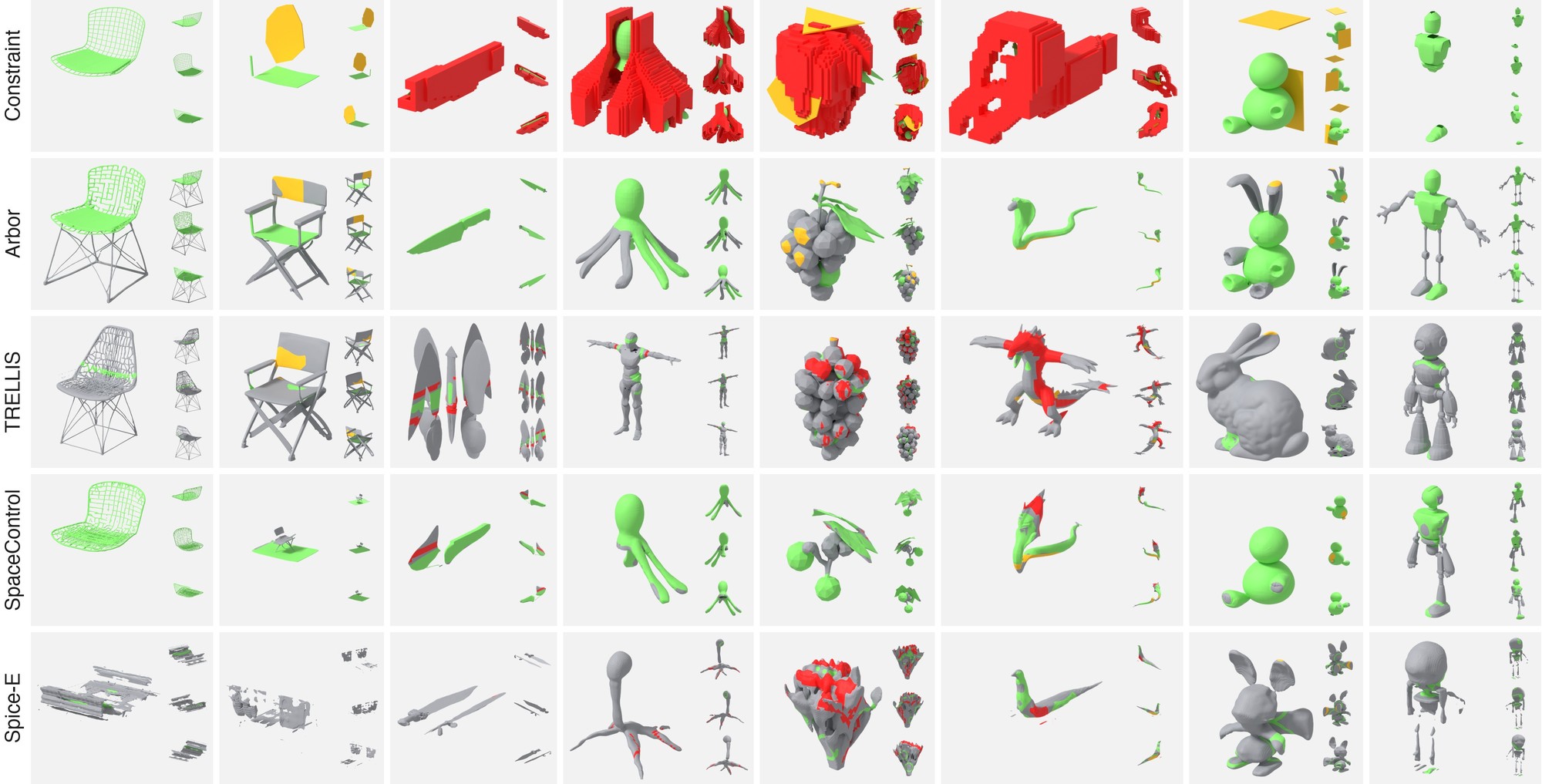}
    \caption{\textbf{Comparison with the baselines on the automatic split.}
    Layout and colors as in Fig.~\ref{fig:comparison}.
    All methods in a column use the same constraint, prompt and seed.}
    \label{fig:supp:comparison_auto}
\end{figure*}

\begin{figure*}[!tbp]
    \centering
    \includegraphics[width=\textwidth]{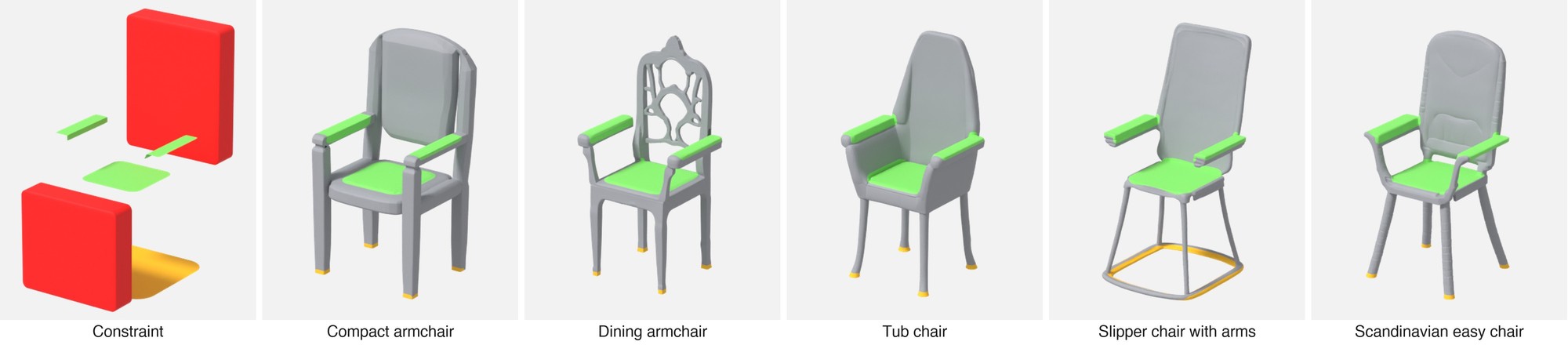}
    \caption{\textbf{One constraint with different prompts.}
    The constraint of Fig.~\ref{fig:introduction:overview} with further prompts, which are given below each object.
    Colors as in Fig.~\ref{fig:comparison}.}
    \label{fig:supp:many_prompts}
\end{figure*}

\par\medskip\noindent\begin{minipage}{\columnwidth}
    \centering
    \footnotesize
    \renewcommand{\arraystretch}{1.05}
    \setlength{\tabcolsep}{4pt}
    \begin{tabular*}{\columnwidth}{@{\extracolsep{\fill}}lcccl@{}}
        \toprule
        Split & Objects & \makecell{With\\avoidance} & \makecell{With\\touch} & Constraints \\
        \midrule
        Automatic & $128$ & $64$ & $63$ & sampled \\
        Manual    & $32$  & $12$ & $7$  & by hand \\
        \bottomrule
    \end{tabular*}
    \captionof{table}{\textbf{The two benchmark splits.}
    Every object has a hull, and the two middle columns count the objects that also have avoidance or touch regions.}
    \label{tab:supp:splits}
\end{minipage}\par\medskip

\begin{figure*}[t]
    \centering
    \setlength{\tabcolsep}{0pt}
    \begin{tabular}{@{}c@{}}
    \includegraphics[width=\textwidth]{figures/assets/supp_sweeps/chair_b.jpg}\\[-1pt]
    {\footnotesize\textbf{Chair}, prompt ``Chair''. A straight chair becomes a lounge chair.}\\[3pt]
    \includegraphics[width=\textwidth]{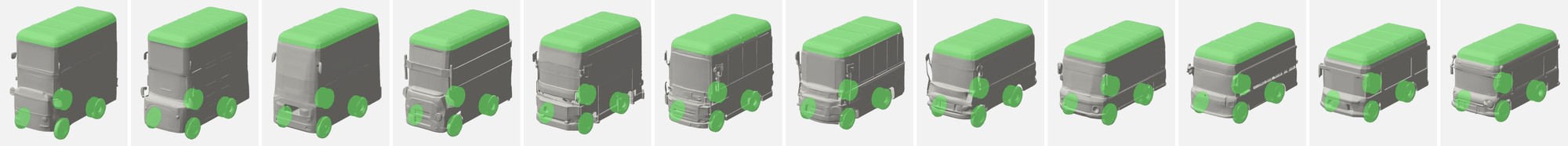}\\[-1pt]
    {\footnotesize\textbf{Bus}, prompt ``A bus''. The roof comes down.}\\[3pt]
    \includegraphics[width=\textwidth]{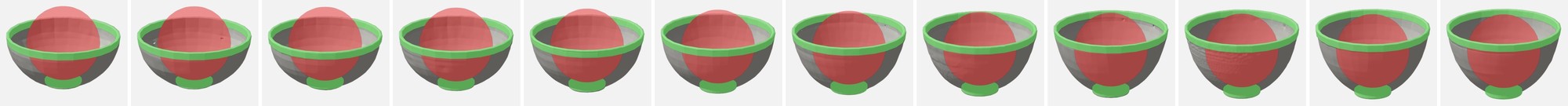}\\[-1pt]
    {\footnotesize\textbf{Bowl}, prompt ``Bowl''. The bowl deepens around a sphere that has to stay free.}\\[3pt]
    \includegraphics[width=\textwidth]{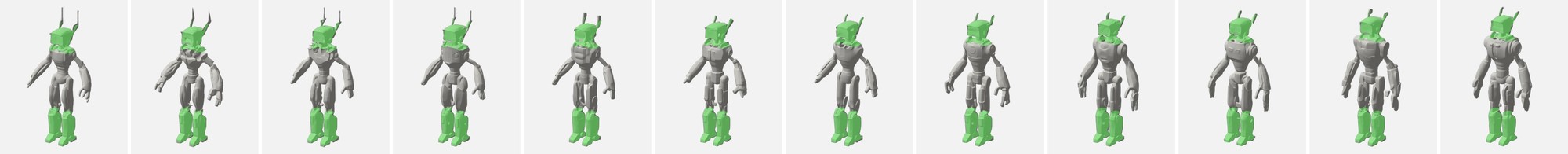}\\[-1pt]
    {\footnotesize\textbf{Robot}, prompt ``A robot.'' Only the feet and the head are given. The head moves up, and the torso and the arms grow with it.}\\[3pt]
    \includegraphics[width=\textwidth]{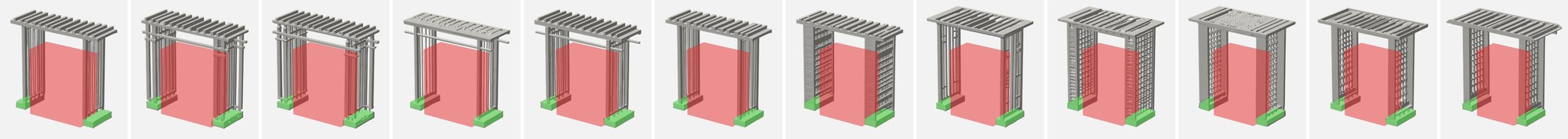}\\[-1pt]
    {\footnotesize\textbf{Garden arbor}, prompted as a wooden trellis arbor. The passage narrows and lattice fills the sides.}
    \end{tabular}
    \caption{\textbf{Constraint sweeps.}
    Each row keeps one prompt and changes the constraint over twelve steps.
    The hull is drawn in green and the avoidance region in red.}
    \label{fig:supp:additional_sweeps}
\end{figure*}

\paragraph{Benchmark.}
We evaluate on Toys4K~\citep{toys4k2021}, which Arbor never sees in training and which TRELLIS also holds out for evaluation~\citep{trellis2024}.

We use two splits of this dataset (Tab.~\ref{tab:supp:splits}).
For the automatic split we take $128$ objects, draw one constraint for each of them with the constraint sampler, and keep it fixed.
For the manual split we modeled $32$ constraint sets by hand in Blender, one object at a time, and we show all of them in Fig.~\ref{fig:supp:manual_sets}.
Many of these sets give only a few parts of their object, such as the seat of a chair, the wheels of a bicycle or the taps of a sink.
Several sets also add a block of space that has to stay empty or a plane that the object has to touch.
The manual sets do not come from the training sampler, so this split tests whether Arbor follows what a person intends.

\subsection{Additional Qualitative Results}\label{app:extended_results}

\suppsummary{In this section we show more objects that Arbor generates. We show sixteen further objects, compare Arbor with the baselines on the automatic split, combine one constraint with different prompts, and add five constraint sweeps.}

\paragraph{Gallery.}
In Fig.~\ref{fig:supp:gallery2} we show sixteen further objects that do not appear in the main paper.
For the first half we built new constraints from boxes, cylinders and ellipsoids, without a reference object.
Each of these constraints gives only a few parts of the object, such as the posts of a bed, the sails of a windmill or the handle and the spout of a watering can.
Arbor then generates the rest of the object, for example the mattress, the tower of the windmill and the body of the can.

For the second half we use sets of the automatic split, which are cut from an object and therefore carry more detail.
In both cases Arbor completes the object of the prompt around the given parts.

\begin{table*}[!tbp]
    \centering
    \footnotesize
    \renewcommand{\arraystretch}{1.1}
    \setlength{\tabcolsep}{5pt}
    \begin{tabular}{@{}cl p{0.82\textwidth}@{}}
        \toprule
        Col. & Object & Prompt \\
        \midrule
        1 & Chair    & A traditional dark brown wooden rocking chair with curved rockers, turned front legs, simpler back legs, a flat oval seat, curved armrests with supports, a backrest with five tapered vertical spindles, polished finish, shows fine woodworking craftsmanship. \\
        2 & Car      & A low-poly car model in turquoise blue with a boxy, angular design. Flat roof sloping rearward, light gray trapezoidal side windows, rectangular front and rear windows, dark gray tires, light gray hexagonal hubs, and no additional details. \\
        3 & Bus      & A classic red double-decker bus with rounded edges, a horizontal green stripe, black rectangular windows, two black windshields per deck, three pairs of grey wheels, a rear entry with stairs, and small silver side mirrors. \\
        4 & Sink     & A modern, rectangular, white sink with an integrated basin and faucet. The basin has rounded corners, a metallic drain cover, and a single-handle, chrome faucet on the right. The base is a smooth, white rectangular prism. \\
        5 & Sofa     & Two-seater pastel pink fabric-upholstered sofa with rectangular backrest, tall back cushions divided into two segments, rounded broad armrests with faint button detailing, and short dark legs. Modern minimalist design suitable for residential use. \\
        6 & Airplane & A stylized airplane with a cylindrical fuselage tapering at the rear, two large wings, and a tail with vertical and horizontal stabilizers. The white and light pink fuselage has cyan-blue windows. Engines under wings are pink with darker intakes. \\
        7 & Sailboat & A minimally styled sailboat with a V-shaped hull, flat rectangular deck, mid-deck rectangular cabin with a grey roof, light beige mast, white and light reddish-pink sails, and light beige booms, rudder, and rigging. \\
        8 & Piano    & A small red toy piano with a rounded rectangular body, short stubby legs, simple keyboard front, playful proportions, glossy painted finish, and compact childlike musical-instrument design. \\
        \bottomrule
    \end{tabular}
    \caption{\textbf{Prompts of Fig.~\ref{fig:comparison}}, by column from left to right.}
    \label{tab:supp:prompts}
\end{table*}

\paragraph{Baselines on the automatic split.}
The comparison in Fig.~\ref{fig:comparison} uses hand authored constraints.
In Fig.~\ref{fig:supp:comparison_auto} we repeat it on eight sets of the automatic split, where several hulls are surrounded by large avoidance regions.
TRELLIS does not receive the constraint and therefore generates a different object, for example a standing figure for the octopus and a dragon for the cobra.
SpaceControl stays close to the hull, so the wire chair has no legs and the bunny has no ears.
Spice-E breaks apart or enters the avoidance regions.
Arbor follows the regions and still completes the object.
The numbers for this split are given in Tab.~\ref{tab:metrics}.

\paragraph{One constraint with different prompts.}
In the introduction we consider a set of chairs that all share one sitting animation.
To show this with more objects, we take the constraint of the teaser and combine it with further prompts (Fig.~\ref{fig:supp:many_prompts}).
The two regions that have to stay empty remain empty for every prompt, and the back, the legs and the style follow the prompt.

\paragraph{Sweeps.}
In a sweep we keep the prompt fixed and change the constraint in small steps.
We built the constraints of each sweep for this purpose, and where they start from a source object they move away from its geometry.
In Fig.~\ref{fig:supp:additional_sweeps} we show five sweeps over twelve steps that are not in the main paper.
A straight chair leans back into a lounge chair, a bus follows its roof down, and a bowl deepens around a sphere that has to stay free.
For the robot we give only the feet and the head, which we cut from a Toys4K object, and as the head moves up Arbor builds a longer torso and longer arms between them.
The garden arbor has no source object.
Here we give only two foot pads and a passage that has to stay free, and as the passage gets narrower the pads move inward and Arbor fills the sides with lattice.

The videos in the folder \texttt{sweeps/} play these five sweeps and the sweeps of Fig.~\ref{fig:constraint_sweep} in motion.

\paragraph{Prompts of Fig.~\ref{fig:comparison}.}\label{app:fig3_prompts}
Tab.~\ref{tab:supp:prompts} lists the full prompts of Fig.~\ref{fig:comparison}.
Every method receives the same prompt, constraint and seed.

\renewcommand{\dblfloatpagefraction}{0.99}
\subsection{Additional Quantitative Results}\label{app:more_numbers}\label{app:variation}

\suppsummary{In this section we test how reliable the comparison of Tab.~\ref{tab:metrics} is. We add confidence intervals, give Arbor the same input as the baselines, change the form of the score, group the results by constraint type and measure the decoded surface. Arbor stays ahead of every baseline in each of these tests.}

\paragraph{Intervals and paired differences.}
In Tab.~\ref{tab:ctrl_ci} we add $95\%$ bootstrap intervals to the Ctrl column of Tab.~\ref{tab:metrics}.
Some objects are easy and others are hard for every method, so we also compute the paired difference in Ctrl between Arbor and each method on the same objects.
All paired intervals between Arbor and the baselines lie above zero on both splits.
We also repeated the evaluation of Arbor with four seeds.
Ctrl is $0.780 \pm 0.011$ on the manual split and $0.926 \pm 0.003$ on the automatic split, so the seed changes the score far less than the choice of method.
Arbor with the TRELLIS decoder scores practically the same in Ctrl, because the decoder adapters act below the $64^3$ grid on which the terms of Ctrl are measured.

\begin{table}[t]
    \centering
    \scriptsize
    \renewcommand{\arraystretch}{1.05}
    \setlength{\tabcolsep}{2pt}
    \newcommand{\gt}[1]{\textcolor{black!50}{#1}}
    \begin{tabular*}{\columnwidth}{@{\extracolsep{\fill}}l ccc@{}}
        \toprule
        Method & Ctrl & Arbor $-$ method & F \\
        \midrule
        \multicolumn{4}{@{}l}{\emph{Manual split, $n = 32$}} \\
        TRELLIS      & .403 [.312, .492] & $+.360$ [$+.281$, $+.445$] & .243 \\
        Gradient     & .596 [.503, .677] & $+.167$ [$+.076$, $+.251$] & .346 \\
        SpaceControl & .487 [.394, .580] & $+.276$ [$+.160$, $+.391$] & .479 \\
        Spice-E      & .202 [.140, .265] & $+.561$ [$+.461$, $+.657$] & .180 \\
        \textbf{Arbor (ours)} & .763 [.668, .847] & -- & .511 \\
        Arbor, TRELLIS decoder & .766 [.672, .850] & $-.003$ [$-.009$, $+.001$] & .509 \\
        \gt{GT latent, decoded} & \gt{.895 [.810, .959]} & \gt{--} & \gt{.782} \\
        \midrule
        \multicolumn{4}{@{}l}{\emph{Automatic split, $n = 128$}} \\
        TRELLIS      & .436 [.399, .474] & $+.488$ [$+.444$, $+.530$] & .273 \\
        Gradient     & .697 [.668, .725] & $+.228$ [$+.193$, $+.260$] & .458 \\
        SpaceControl & .631 [.586, .675] & $+.293$ [$+.252$, $+.337$] & .584 \\
        Spice-E      & .321 [.285, .358] & $+.604$ [$+.563$, $+.643$] & .242 \\
        \textbf{Arbor (ours)} & .925 [.902, .942] & -- & .765 \\
        Arbor, TRELLIS decoder & .925 [.902, .942] & $+.000$ [$-.002$, $+.002$] & .763 \\
        \gt{GT latent, decoded} & \gt{.958 [.944, .970]} & \gt{--} & \gt{.895} \\
        \bottomrule
    \end{tabular*}
    \caption{\textbf{Intervals and paired differences for Tab.~\ref{tab:metrics}.}
    Ctrl with $95\%$ bootstrap intervals over objects, and the paired difference in Ctrl between Arbor and each row on the same objects.
    F is the F-score to the source object, which is not a term of Ctrl.
    The row with the TRELLIS decoder is Arbor without the decoder adapters.}
    \label{tab:ctrl_ci}
\end{table}

\paragraph{Arbor with the input of the baselines.}
None of the existing methods has an input for avoidance or touch regions, so we give the baselines the merged hull.
This means that Arbor receives more information than the baselines, and its lead could simply come from this richer input.
To test this, we remove the avoidance and touch regions from the input of Arbor and leave the model and all settings unchanged.
Even with the hull alone, Arbor stays ahead of every baseline on both splits, and the paired differences lie above zero (Tab.~\ref{tab:hull_only}).

\begin{table}[t]
    \centering
    \scriptsize
    \renewcommand{\arraystretch}{1.05}
    \setlength{\tabcolsep}{2pt}
    \begin{tabular*}{\columnwidth}{@{\extracolsep{\fill}}l c c c@{}}
        \toprule
        Method & Ctrl & \makecell{Arbor, hull only\\$-$ method} & \makecell{Avoid\\resp.} \\
        \midrule
        \multicolumn{4}{@{}l}{\emph{Manual split}} \\
        TRELLIS      & 0.403 & $+0.328$ [$+0.252$, $+0.411$] & 0.667 \\
        Gradient     & 0.596 & $+0.134$ [$+0.048$, $+0.216$] & 0.500 \\
        SpaceControl & 0.487 & $+0.243$ [$+0.126$, $+0.360$] & 0.917 \\
        Spice-E      & 0.202 & $+0.529$ [$+0.423$, $+0.628$] & 0.417 \\
        Arbor, hull only  & 0.731 & -- & 0.667 \\
        \textbf{Arbor, all inputs} & \textbf{0.763} & -- & \textbf{1.000} \\
        \midrule
        \multicolumn{4}{@{}l}{\emph{Automatic split}} \\
        TRELLIS      & 0.436 & $+0.460$ [$+0.417$, $+0.503$] & 0.266 \\
        Gradient     & 0.697 & $+0.199$ [$+0.167$, $+0.231$] & 0.281 \\
        SpaceControl & 0.631 & $+0.265$ [$+0.220$, $+0.313$] & 0.641 \\
        Spice-E      & 0.321 & $+0.575$ [$+0.534$, $+0.616$] & 0.312 \\
        Arbor, hull only  & 0.896 & -- & 0.578 \\
        \textbf{Arbor, all inputs} & \textbf{0.925} & -- & \textbf{0.969} \\
        \bottomrule
    \end{tabular*}
    \caption{\textbf{Arbor with the input of the baselines.}
    The hull only run of Arbor receives the merged hull and the prompt, without avoidance and touch regions.
    The middle column is the paired difference in Ctrl between this run and each baseline, with $95\%$ bootstrap intervals.
    Avoid resp.\ is the share of objects whose avoidance regions are fully respected.
    Without the avoidance input Arbor cannot know which regions have to stay empty, which explains the lower Avoid resp.\ of this run.}
    \label{tab:hull_only}
\end{table}

\paragraph{Variants of the control score.}
The ranking of the methods should not depend on how exactly we build Ctrl.
We therefore recompute Ctrl with three other forms of the avoidance term (Tab.~\ref{tab:ctrl_sensitivity}).
We also remove the volume term, which is the only term that looks at the source object, and then the prompt term (Tab.~\ref{tab:ctrl_terms}).
Arbor ranks first in every column on both splits, and its paired difference to every baseline stays above zero without the volume term and without the prompt term.
This shows that the lead of Arbor does not come from similarity to the source object.

\begin{table}[t]
    \centering
    \scriptsize
    \setlength{\tabcolsep}{3pt}
    \begin{tabular*}{\columnwidth}{@{\extracolsep{\fill}}l cccc cccc@{}}
        \toprule
        & \multicolumn{4}{c}{Manual} & \multicolumn{4}{c}{Auto} \\
        \cmidrule(lr){2-5} \cmidrule(l){6-9}
        Method & Depth & Bin. & Pen. & Vox. & Depth & Bin. & Pen. & Vox. \\
        \midrule
        TRELLIS      & .403 & .380 & .425 & .424 & .436 & .271 & .454 & .453 \\
        Gradient     & .596 & .476 & .624 & .617 & .697 & .482 & .734 & .732 \\
        SpaceControl & .487 & .487 & .487 & .487 & .631 & .527 & .643 & .643 \\
        Spice-E      & .202 & .164 & .220 & .220 & .321 & .214 & .336 & .336 \\
        \textbf{Arbor (ours)} & \textbf{.763} & \textbf{.763} & \textbf{.763} & \textbf{.763} & \textbf{.925} & \textbf{.911} & \textbf{.925} & \textbf{.925} \\
        \bottomrule
    \end{tabular*}
    \caption{\textbf{Ctrl under other avoidance terms.}
    Depth is the term of the paper.
    Bin.\ replaces it by one if the regions are respected and zero otherwise, Pen.\ by one minus the share of the surface inside the region, and Vox.\ by one minus the overlap of surface voxels with the region.}
    \label{tab:ctrl_sensitivity}
\end{table}

\begin{table}[t]
    \centering
    \small
    \setlength{\tabcolsep}{3pt}
    \newcommand{\gt}[1]{\textcolor{black!50}{#1}}
    \begin{tabular*}{\columnwidth}{@{\extracolsep{\fill}}l ccc ccc@{}}
        \toprule
        & \multicolumn{3}{c}{Manual} & \multicolumn{3}{c}{Auto} \\
        \cmidrule(lr){2-4} \cmidrule(l){5-7}
        Method & Ctrl & \makecell{no\\Vol.} & \makecell{no\\PR} & Ctrl & \makecell{no\\Vol.} & \makecell{no\\PR} \\
        \midrule
        TRELLIS      & .403 & .404 & .341 & .436 & .404 & .372 \\
        Gradient     & .596 & .723 & .561 & .697 & .762 & .672 \\
        SpaceControl & .487 & .755 & .465 & .631 & .758 & .603 \\
        Spice-E      & .202 & .181 & .168 & .321 & .290 & .281 \\
        \textbf{Arbor (ours)} & \textbf{.763} & \textbf{.904} & \textbf{.741} & \textbf{.925} & \textbf{.947} & \textbf{.928} \\
        \midrule
        \gt{GT latent, decoded} & \gt{.895} & \gt{.937} & \gt{.888} & \gt{.958} & \gt{.950} & \gt{.964} \\
        \bottomrule
    \end{tabular*}
    \caption{\textbf{Ctrl with one term removed.}
    The score of the paper next to the same harmonic mean without the volume term and without the prompt term.
    The paired difference between Arbor and every baseline stays above zero in both variants on both splits.}
    \label{tab:ctrl_terms}
\end{table}

\paragraph{Constraint types.}
A constraint can consist of a hull alone, or it can add avoidance regions, touch regions or both.
In Tab.~\ref{tab:ctrl_by_type} we group the objects of the automatic split by these four cases.
Arbor leads every baseline in each group, including the objects that carry a hull and nothing else.

\begin{table}[t]
    \centering
    \footnotesize
    \setlength{\tabcolsep}{2pt}
    \begin{tabular*}{\columnwidth}{@{\extracolsep{\fill}}l c c cccc@{}}
        \toprule
        & & & \multicolumn{4}{c}{Arbor $-$ method} \\
        \cmidrule(l){4-7}
        Types & $n$ & Arbor & TRELLIS & Grad. & SpaceC. & Spice-E \\
        \midrule
        Hull              & 30 & .897 & $+.485$ & $+.170$ & $+.282$ & $+.567$ \\
        Hull, avoidance   & 35 & .952 & $+.459$ & $+.324$ & $+.283$ & $+.633$ \\
        Hull, touch       & 34 & .930 & $+.510$ & $+.173$ & $+.341$ & $+.589$ \\
        All three         & 29 & .914 & $+.501$ & $+.236$ & $+.262$ & $+.626$ \\
        \bottomrule
    \end{tabular*}
    \caption{\textbf{Ctrl by constraint type on the automatic split.}
    Objects are grouped by the types of region their constraint contains.
    The right part gives the paired difference in Ctrl between Arbor and each baseline.
    All $95\%$ bootstrap intervals of these differences lie above zero.}
    \label{tab:ctrl_by_type}
\end{table}

\paragraph{Detail at the decoded surface.}
We measure the terms of Ctrl on a $64^3$ grid, which does not show how closely the final surface follows the hull.
In Tab.~\ref{tab:detail256_supp} we therefore measure the decoded surface at $256^3$ for all methods.
Arbor reaches more of the hull surface than every baseline on both splits and has the lowest hull deviation.
The two Arbor rows decode the same latents, so the difference between them is the effect of the decoder adapters alone.
We show this effect on three objects in Fig.~\ref{fig:supp:decoder_more}.
The TRELLIS decoder leaves the surface next to the hull, and the hull decoder brings it onto the hull.
With the hull decoder Arbor reaches about the level of ground truth latents decoded by TRELLIS.

\begin{table}[t]
    \centering
    \scriptsize
    \renewcommand{\arraystretch}{1.0}
    \setlength{\tabcolsep}{4pt}
    \newcommand{\gt}[1]{\textcolor{black!50}{#1}}
    \newcommand{\sw}[1]{\textcolor{#1}{\rule{0.55em}{0.55em}}\,}
    \begin{tabular*}{\columnwidth}{@{\extracolsep{\fill}}lcccc@{}}
        \toprule
        & \multicolumn{2}{c}{\sw{hull}Hull at $256^3$ $\uparrow$} & \multicolumn{2}{c}{\sw{hull}Hull deviation $\downarrow$} \\
        \cmidrule(lr){2-3} \cmidrule(l){4-5}
        Method & Manual & Auto & Manual & Auto \\
        \midrule
        TRELLIS                 & 0.108 & 0.098 & 10.49 & 13.86 \\
        Gradient                & 0.238 & 0.273 & 2.56 & 1.84 \\
        SpaceControl            & 0.502 & 0.447 & 1.53 & 3.32 \\
        Spice-E                 & 0.025 & 0.050 & 19.42 & 15.41 \\
        Arbor, TRELLIS decoder  & 0.554 & 0.662 & 1.28 & 0.91 \\
        \textbf{Arbor, hull decoder} & \textbf{0.699} & \textbf{0.774} & \textbf{0.94} & \textbf{0.69} \\
        \midrule
        \gt{GT latent, TRELLIS decoder} & \gt{0.714} & \gt{0.774} & \gt{0.95} & \gt{0.81} \\
        \gt{GT latent, hull decoder}    & \gt{0.779} & \gt{0.816} & \gt{0.81} & \gt{0.70} \\
        \gt{Source mesh}                & \gt{0.979} & \gt{0.966} & \gt{0.07} & \gt{0.05} \\
        \bottomrule
    \end{tabular*}
    \caption{\textbf{Hull detail at $256^3$, extended.}
    Tab.~\ref{tab:detail256} with all baselines and the source mesh.
    Hull deviation is the mean distance from the hull surface to the generated surface in $256^3$ voxels.
    The paired difference in hull overlap between Arbor with the hull decoder and every baseline lies above zero on both splits.}
    \label{tab:detail256_supp}
\end{table}

\begin{figure*}[!tbp]
    \centering
    \includegraphics[width=\textwidth]{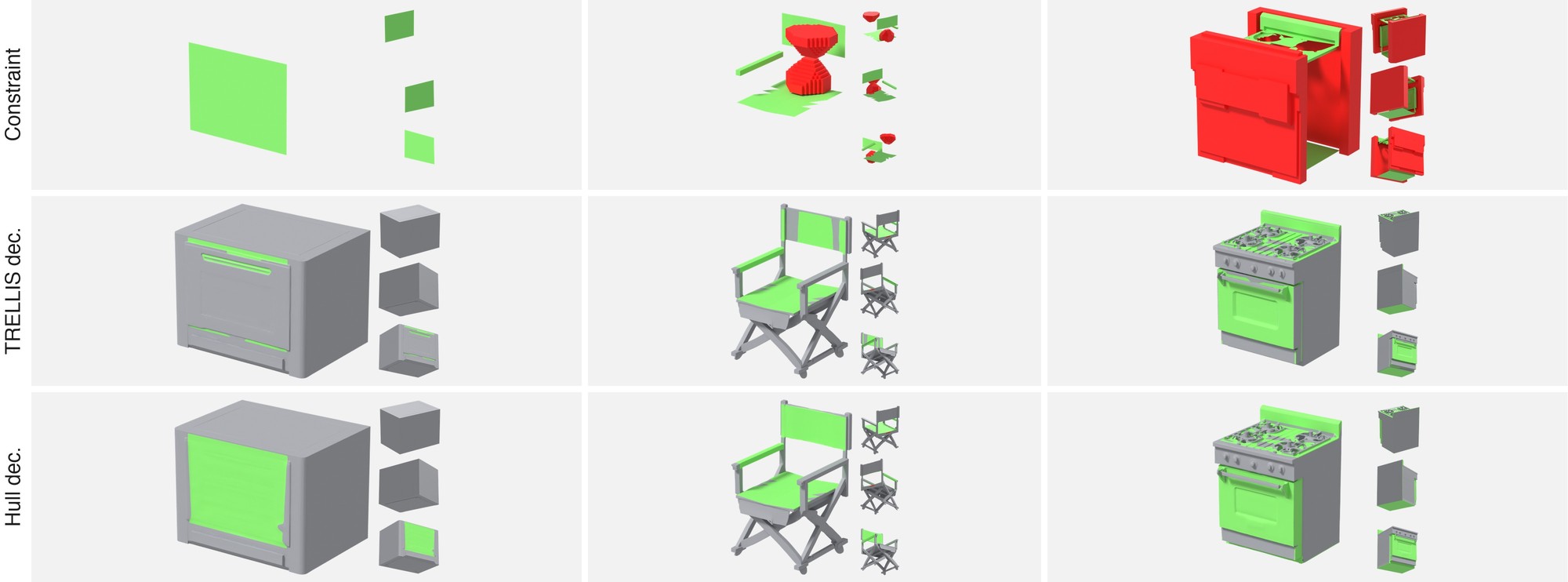}
    \caption{\textbf{Hull decoder and TRELLIS decoder on the same latents.}
    The TRELLIS decoder sets the screen of the monitor back behind the hull and leaves parts of the backrest of the chair off it.
    The hull decoder places both surfaces on the hull.
    Colors as in Fig.~\ref{fig:comparison}.}
    \label{fig:supp:decoder_more}
\end{figure*}

\paragraph{Conditioning probes.}
In the main paper we remove parts of the constraint tokens at test time to see what the trained model relies on (Tab.~\ref{tab:probe_findings}).
In Tab.~\ref{tab:probes_auto} we repeat these probes on the automatic split, which has four times as many objects, and we find the same result.
Without positions the hull overlap falls far below that of the generator without geometry, and without the role of a region it falls close to it.
In Fig.~\ref{fig:supp:probes} we show the probes on constraints of the manual split.

Without positions the object falls apart into scattered pieces, and without roles the truck fills the region that has to stay empty.

\begin{table}[t]
    \centering
    \scriptsize
    \renewcommand{\arraystretch}{1.0}
    \setlength{\tabcolsep}{0pt}
    \newcommand{\sw}[1]{\textcolor{#1}{\rule{0.55em}{0.55em}}\,}
    \begin{tabular*}{\columnwidth}{@{\extracolsep{\fill}}lcccccc@{}}
        \toprule
        & \textbf{Full} & \makecell{Local\\only} & \makecell{No\\shape} & \makecell{No\\role} & \makecell{No\\pos.\ enc.} & \makecell{No\\geometry} \\
        \midrule
        \multicolumn{7}{@{}l}{\emph{Manual split, $64$ samples per probe (Tab.~\ref{tab:probe_findings})}} \\
        \sw{hull}Hull$\uparrow$          & \textbf{.896} & .715 & .715 & .274 & .051 & .246 \\
        \sw{avoidance}Avoid$\uparrow$ & \textbf{.981} & .813 & .918 & .819 & .562 & .777 \\
        Vol.$\uparrow$ & \textbf{.717} & .254 & .666 & .678 & .293 & .586 \\
        \midrule
        \multicolumn{7}{@{}l}{\emph{Automatic split, $256$ samples per probe}} \\
        \sw{hull}Hull$\uparrow$          & \textbf{.950} & .834 & .783 & .311 & .035 & .248 \\
        \sw{avoidance}Avoid$\uparrow$ & \textbf{.999} & .953 & .918 & .711 & .551 & .695 \\
        Vol.$\uparrow$ & \textbf{.899} & .294 & .813 & .692 & .351 & .677 \\
        \bottomrule
    \end{tabular*}
    \caption{\textbf{Conditioning probes on both splits.}
    Parts of the constraint tokens are removed at test time, with two seeds per object.
    The upper block repeats Tab.~\ref{tab:probe_findings}, and the lower block shows the same probes on the automatic split.}
    \label{tab:probes_auto}
\end{table}

\begin{figure*}[!tbp]
    \centering
    \includegraphics[width=\textwidth]{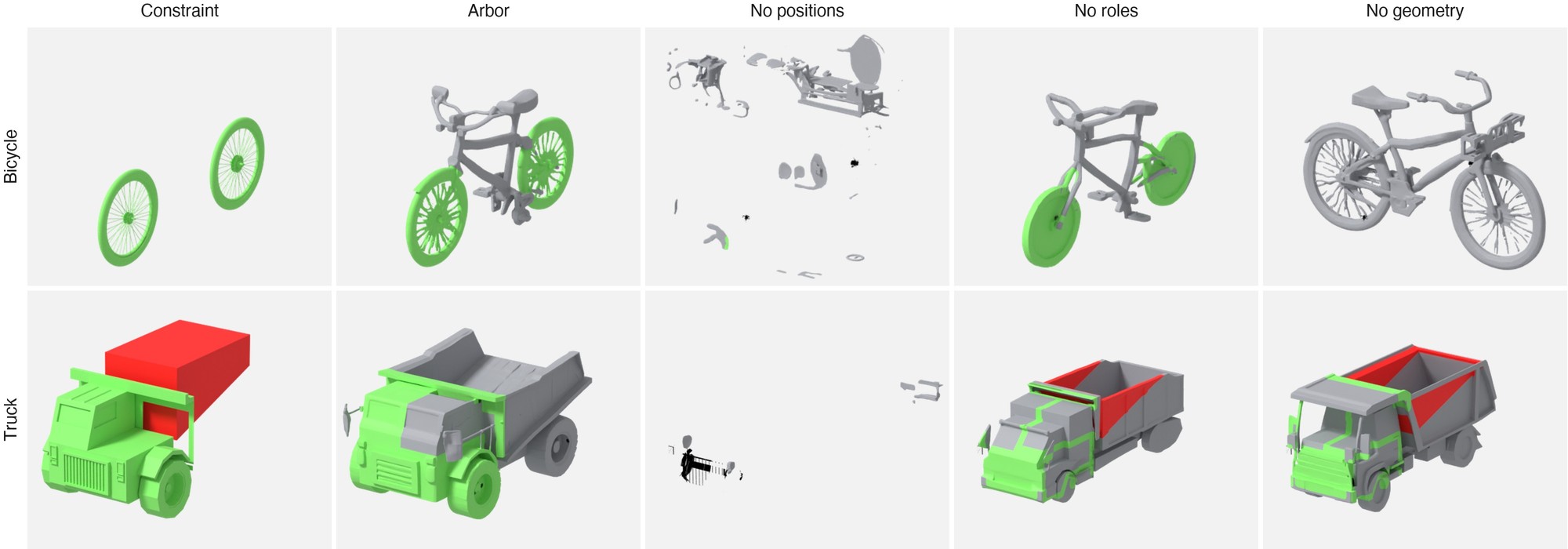}
    \caption{\textbf{Conditioning probes.}
    Each column removes one part of the constraint tokens of the trained model at test time, with the same prompt and seed.
    Colors as in Fig.~\ref{fig:comparison}.}
    \label{fig:supp:probes}
\end{figure*}

\subsection{Design Choices}\label{app:variants}

\suppsummary{In this section we explain the main design choices of Arbor. Guidance at sampling time needs no training, but it leaves rough and torn surfaces. An additional loss on the constraint does not improve Ctrl, so we train Arbor with the plain training loss. We also replace the pretrained tokens with a trained $16^3$ field. This field learns to place the coarse structure as well, but it reaches less of the hull and far less of the decoded surface.}

\begin{figure*}[!tbp]
    \centering
    \includegraphics[width=\textwidth]{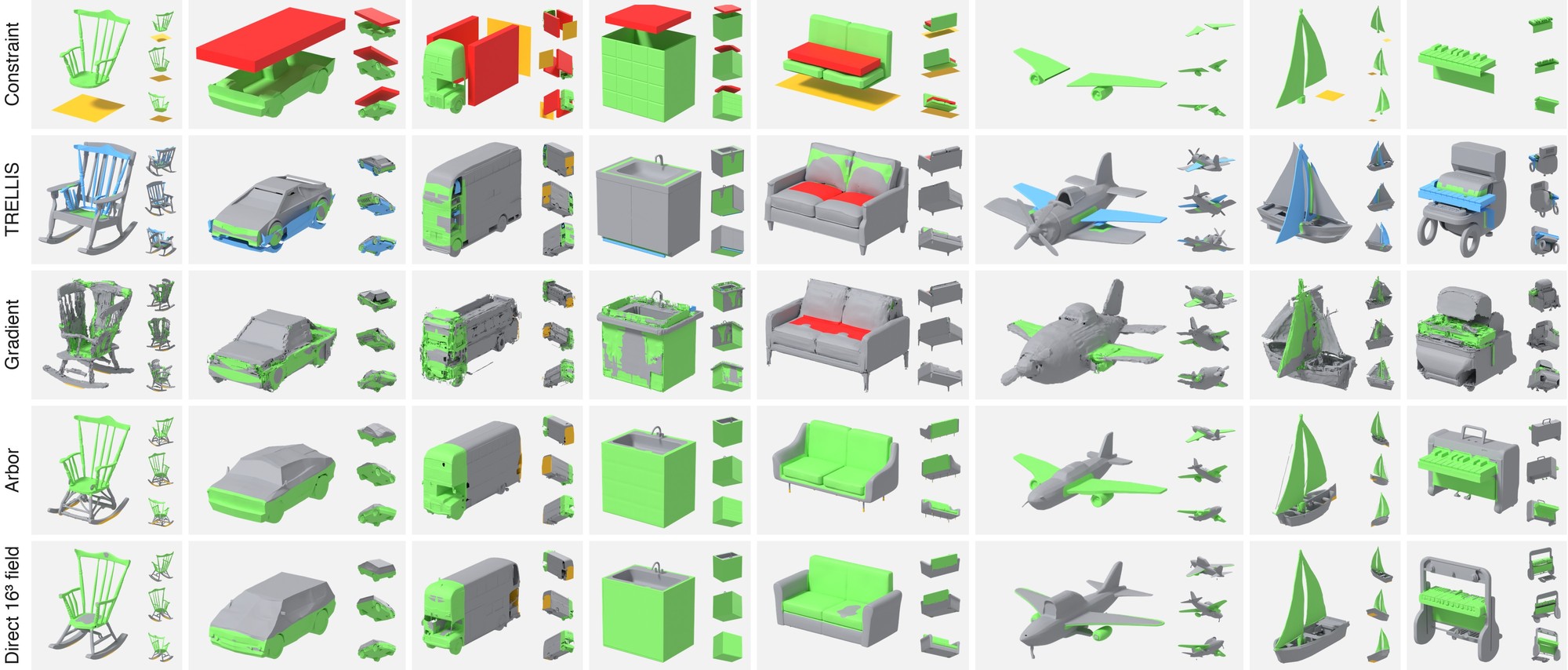}
    \caption{\textbf{Guidance at sampling time and a trained $16^3$ field next to Arbor.}
    The cases and seeds of Fig.~\ref{fig:comparison}.
    Gradient steers TRELLIS at sampling time (Sec.~\ref{app:gradient_baseline}) and leaves rough and torn surfaces.
    The direct $16^3$ field is trained for the same $60$k steps as Arbor and decoded with the TRELLIS decoder.
    It places the coarse structure but misses parts of the hull, such as the wings of the plane and parts of the seat of the chair.
    Colors as in Fig.~\ref{fig:comparison}.}
    \label{fig:supp:encoders}
\end{figure*}

\paragraph{Guidance at sampling time.}\label{app:gradient_baseline}
One way to control a frozen generator without any training is to steer it while it samples.
Our Gradient baseline does this on TRELLIS with the gradient of a hull loss.
At a guided step~$k$ we decode the current sparse structure latent~$x_k$ into occupancy logits on the $64^3$ grid.
The loss~$\mathcal{L}_{\mathrm{hull}}$ is the binary cross entropy between these logits and one, averaged over the voxels of the hull, so it penalizes hull voxels that are still empty.
We then move the latent against the normalized gradient,
\[
\begin{aligned}
g_k &= \nabla_{x_k}\mathcal{L}_{\mathrm{hull}}(x_k), \\
x_k &\leftarrow x_k - \eta_k \bigl(1 + 10\,\mathcal{L}_{\mathrm{hull}}(x_k)\bigr) \frac{g_k}{\lVert g_k \rVert_2}.
\end{aligned}
\]
We apply this guidance during the last $14$ of the $25$ sampling steps and let $\eta_k$ grow linearly from zero to $0.2$ over these steps.
The text guidance is the same as in TRELLIS.

Gradient pulls mass into the hull and is the strongest baseline in Ctrl (Tab.~\ref{tab:metrics}).
The decoded objects, however, have rough and torn surfaces (Fig.~\ref{fig:supp:encoders}).
Where the constraint asks for a different shape, the unguided object remains and the hull is simply added to it.
We believe that the reason is the loss, which only asks for occupancy inside the hull and does not know which object this occupancy should belong to.
When we give Arbor the same input, the hull alone, it stays ahead of Gradient on both splits (Tab.~\ref{tab:hull_only}).

\paragraph{Encoder.}
The constraint has to reach the generator in a form that it can read.
In Arbor we use the frozen encoders of TRELLIS.2~\citep{trellis22025}, because their tokens keep the detail of the constraint.
The control can also be learned with other mechanisms.
As an example, we replace the constraint tokens and the router with a direct field on the $16^3$ grid of the generator and train it together with the adapters for the same $60$k steps as Arbor.
The trained $16^3$ field also learns to place the coarse structure, but it reaches less of the hull.
The structure stage is coarse, however, and our goal is to control the mesh and not only the sparse structure.
The tokens of Arbor act in the structure stage and in the decoder.
A $16^3$ field carries no information below its grid, so it fits only the structure stage and cannot feed the decoder adapters, and we decode it with the TRELLIS decoder.
At the decoded surface the difference becomes clear (Tab.~\ref{tab:supp:detail256_enc60k}).
The $16^3$ field reaches far less of the hull surface at $256^3$ than Arbor.
Arbor also reaches more of it when both use the TRELLIS decoder, which shows that its tokens carry the detail of the constraint.
We show the difference on the plane, the piano, the chair and the seat of the sofa in Fig.~\ref{fig:supp:encoders}.

\begin{table}[t]
    \centering
    \scriptsize
    \renewcommand{\arraystretch}{1.0}
    \setlength{\tabcolsep}{4pt}
    \newcommand{\sw}[1]{\textcolor{#1}{\rule{0.55em}{0.55em}}\,}
    \begin{tabular*}{\columnwidth}{@{\extracolsep{\fill}}lcccc@{}}
        \toprule
        & \multicolumn{2}{c}{\sw{hull}Hull at $256^3$ $\uparrow$} & \multicolumn{2}{c}{\sw{hull}Hull deviation $\downarrow$} \\
        \cmidrule(lr){2-3} \cmidrule(l){4-5}
        Method & Manual & Auto & Manual & Auto \\
        \midrule
        Direct $16^3$ field, TRELLIS decoder & 0.420 & 0.536 & 3.22 & 1.25 \\
        Arbor, TRELLIS decoder                & 0.554 & 0.662 & 1.28 & 0.91 \\
        \textbf{Arbor, hull decoder}          & \textbf{0.699} & \textbf{0.774} & \textbf{0.94} & \textbf{0.69} \\
        \bottomrule
    \end{tabular*}
    \caption{\textbf{Hull detail at $256^3$ with the direct $16^3$ field.}
    The measures of Tab.~\ref{tab:detail256}.
    With the same TRELLIS decoder Arbor reaches more of the hull surface than the direct field, and the paired difference lies above zero on both splits.}
    \label{tab:supp:detail256_enc60k}
\end{table}

\paragraph{Training loss.}
We train Arbor with the flow matching loss alone.
This works because every training constraint is cut from the training object, so the model can lower the loss by reading the constraint.
We also tried additional losses that compare the decoded structure directly with the hull, avoidance and touch regions (Tab.~\ref{tab:supp:loss}).
These losses raise hull overlap and avoidance, but prompt retrieval and volume fall, and Ctrl does not improve (it falls on the automatic split).
The model then follows the constraint more closely, but it loses part of the object that the prompt asks for.

\begin{table}[t]
    \centering
    \small
    \setlength{\tabcolsep}{4pt}
    \newcommand{\sw}[1]{\textcolor{#1}{\rule{0.55em}{0.55em}}\,}
    \begin{tabular*}{\columnwidth}{@{\extracolsep{\fill}}lccccc@{}}
        \toprule
        Split & \sw{hull}Hull & \sw{avoidance}Avoid & PR & Vol. & \textbf{Ctrl} \\
        \midrule
        Manual    & $+0.055$ & $+0.060$ & $-0.059$ & $-0.066$ & $-0.032$ \\
        Automatic & $+0.027$ & $+0.003$ & $-0.042$ & $-0.143$ & $-0.101$ \\
        \bottomrule
    \end{tabular*}
    \caption{\textbf{Adding a loss on the constraint.}
    Change of each term when the training of a smaller variant of Arbor is continued with additional losses on the decoded hull, avoidance and touch regions.
    The losses raise hull overlap and avoidance but cost prompt retrieval and volume, and Ctrl does not improve.}
    \label{tab:supp:loss}
\end{table}

\paragraph{Constraint types.}
We focus on hull and avoidance in this paper, because they state where the object has to be and where it has to stay away.
Touch is a looser form of control, which asks the object to reach a surface without saying how.
We include it to show that the same recipe carries over to a further type.
The mechanism itself is not tied to these three types.
For the model a type is a code in the signal channels of the encoder, and each type can cover any number of regions.
A further type therefore needs a new code and training examples for it.

\paragraph{Beyond TRELLIS.}
In this paper we build Arbor into TRELLIS, but its parts are not defined for one particular latent space.
What Arbor needs are geometry tokens with positions, a router that collects them per region, and an adapter in a transformer block.
We already use the same attachment in two networks with different latents, namely the structure denoiser on a dense $16^3$ grid and the mesh decoder on sparse voxels up to $256^3$.
We take this as a sign that the pattern carries over to other latent 3D generators that are built from transformer blocks.
The generator itself is not changed.
All weights of TRELLIS stay identical to the released ones and the adapters start at zero, so Arbor without a constraint is the original generator.

\subsection{Router Capacity and Cost}\label{app:router_k}

\suppsummary{In this section we test whether the token cap of the router limits control, and we measure what Arbor costs at inference. The cap does not reduce control, and it keeps time and memory nearly flat when the constraints grow. If we build the router context once per object, Arbor adds about half a second to the structure stage of TRELLIS.}

\paragraph{Capacity.}
The router gives every group of queries at most $K = 2048$ local tokens.
This cap could limit control when a constraint has many tokens.
To test this, we vary~$K$ at test time on the trained model (Tab.~\ref{tab:router_k}).
Hull overlap, avoidance and volume change little between $K = 256$ and no cap at all, so the cap does not reduce control.

\begin{table}[t]
    \centering
    \small
    \setlength{\tabcolsep}{4.5pt}
    \begin{tabular*}{\columnwidth}{@{\extracolsep{\fill}}lcccccc@{}}
        \toprule
        $K$ & 256 & 512 & 1024 & 2048 & 4096 & all \\
        \midrule
        Hull$\uparrow$  & .899 & .892 & .891 & .896 & .894 & .891 \\
        Avoid$\uparrow$ & .947 & 1.000 & .992 & 1.000 & 1.000 & 1.000 \\
        Vol.$\uparrow$  & .689 & .687 & .734 & .703 & .718 & .722 \\
        \bottomrule
    \end{tabular*}
    \caption{\textbf{Size of the local router context.}
    Arbor on the $32$ hand authored constraint sets, with the number $K$ of local tokens per query group varied at test time.
    The last column removes the cap, and the paper uses $K = 2048$.
    Scores are computed on the generated structure, and Avoid over the $12$ sets that contain an avoidance region.}
    \label{tab:router_k}
\end{table}

\paragraph{Cost.}
In the pipeline that produced all results of the paper, Arbor adds about four seconds per object to the structure stage (Tab.~\ref{tab:overhead}).
Most of this time is spent on rebuilding the router context and its attention layout at every sampling step, although both depend only on the constraint.
If we build them once per object, the outputs are identical and only about half a second remains on top of TRELLIS.
Peak memory hardly changes, and the SLAT stage is not affected.
In Tab.~\ref{tab:router_scale} we also measure the cost for constraints of growing size, up to a fully solid constraint that fills the token grid.
Without the cap, time and memory grow steadily with the number of tokens.
With the cap both stay nearly flat, and for the largest constraint the capped model is about three times faster and needs half the peak memory.

\begin{table}[t]
    \centering
    \small
    \setlength{\tabcolsep}{5pt}
    \begin{tabular*}{\columnwidth}{@{\extracolsep{\fill}}lcc@{}}
        \toprule
        Structure stage & Time (s) & Memory (GB) \\
        \midrule
        TRELLIS, no constraint & 2.22 & 8.97 \\
        Arbor, as evaluated in the paper & 6.44 & 9.23 \\
        \textbf{Arbor, router context built once} & \textbf{2.81} & \textbf{9.23} \\
        \bottomrule
    \end{tabular*}
    \caption{\textbf{Cost of the structure stage.}
    Median time and peak memory for sampling and decoding the structure of one object, on the same GPU and over the $32$ hand authored constraint sets.
    All results of the paper come from the pipeline of the second row, which rebuilds the router context at every sampling step.
    Building it once per object, as in the third row, gives identical outputs and leaves about half a second on top of TRELLIS.}
    \label{tab:overhead}
\end{table}

\begin{table}[t]
    \centering
    \small
    \setlength{\tabcolsep}{4pt}
    \begin{tabular*}{\columnwidth}{@{\extracolsep{\fill}}r cc cc@{}}
        \toprule
        & \multicolumn{2}{c}{$K = 2048$} & \multicolumn{2}{c}{No cap} \\
        \cmidrule(lr){2-3} \cmidrule(l){4-5}
        Tokens & Time (s) & Mem.\ (GB) & Time (s) & Mem.\ (GB) \\
        \midrule
        4{,}096  & 7.26 & 8.35 & 8.04 & 8.97 \\
        8{,}192  & 7.36 & 8.44 & 11.05 & 10.32 \\
        16{,}384 & 7.92 & 8.62 & 17.15 & 13.00 \\
        32{,}768 & 8.87 & 8.97 & 28.73 & 18.38 \\
        \bottomrule
    \end{tabular*}
    \caption{\textbf{Cost against the size of the constraint.}
    Median time and peak memory of the structure stage on one GPU for synthetic constraints of growing size, with the router cap of the paper and without a cap.
    The last row is a fully solid constraint, the largest the token grid can hold.}
    \label{tab:router_scale}
\end{table}

\subsection{Implementation Details}\label{app:implementation_details}

\suppsummary{We list all settings of Arbor in Tab.~\ref{tab:supp:settings} and describe the guidance, the model, the training and the decoder adapters. We train only the attachment, and TRELLIS and the encoders of TRELLIS.2 stay frozen.}

\begin{table}[t]
    \centering
    \footnotesize
    \renewcommand{\arraystretch}{1.05}
    \setlength{\tabcolsep}{4pt}
    \begin{tabular*}{\columnwidth}{@{\extracolsep{\fill}}p{0.40\columnwidth}p{0.55\columnwidth}@{}}
        \toprule
        \multicolumn{2}{@{}l}{\emph{Generator}} \\
        Model & TRELLIS-text-xlarge, frozen \\
        Structure denoiser & $28$ blocks, width $1280$ \\
        Latent grid & $16^3$ cells, $8$ channels \\
        \midrule
        \multicolumn{2}{@{}l}{\emph{Constraint tokens and router}} \\
        Voxel grid of the constraint & $512^3$ \\
        Token grid & $32^3$, sparse \\
        Channels per token & $32$ shape and $32$ signal \\
        Query groups & $64$ groups of $4 \times 4 \times 4$ \\
        Local tokens per group & at most $K = 2048$ \\
        Summary tokens & $96$ \\
        \midrule
        \multicolumn{2}{@{}l}{\emph{Training of the structure attachment}} \\
        Steps & $60$k \\
        Batch & $8$ GPUs with $4$ objects each \\
        Optimizer & AdamW, learning rate $10^{-4}$, no weight decay \\
        Precision & half, adaptive gradient clipping \\
        Condition dropout & $0.1$ for text, $0.1$ for geometry \\
        Mesh size & decimated above $50$k faces \\
        \midrule
        \multicolumn{2}{@{}l}{\emph{Sampling}} \\
        Steps & $25$ \\
        Guidance scale & $7.5$ for $t \in [0.5, 0.95]$ \\
        \midrule
        \multicolumn{2}{@{}l}{\emph{Decoder adapters}} \\
        Levels & $64^3$, $128^3$, $256^3$ \\
        Steps & $12$k \\
        Batch & $8$ GPUs with $2$ objects each \\
        Loss weights & depth $10$, TSDF $0.01$, SSIM $0.2$, LPIPS $0.2$ \\
        \bottomrule
    \end{tabular*}
    \caption{\textbf{Settings of Arbor.}}
    \label{tab:supp:settings}
\end{table}

\begin{figure*}[!tbp]
    \centering
    \includegraphics[width=\textwidth]{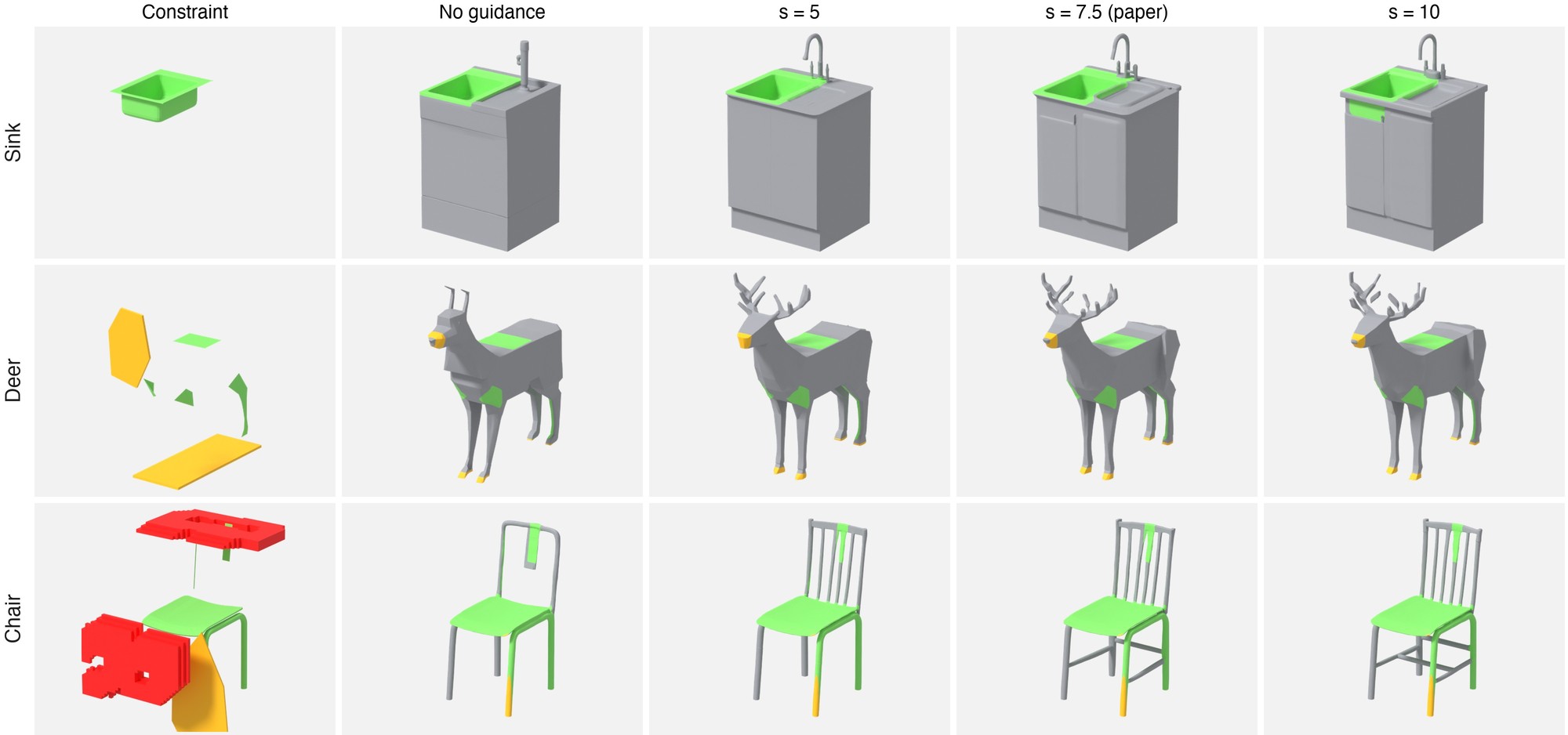}
    \caption{\textbf{Guidance scale with a fixed constraint.}
    Three constraint sets, each generated with one seed at guidance scales from none to $10$.
    Without guidance some parts are lost, such as the tap of the sink and the antlers of the deer.
    With guidance the objects are complete and change little between the scales.
    Colors as in Fig.~\ref{fig:comparison}.}
    \label{fig:supp:guidance}
\end{figure*}

\paragraph{Guidance.}
Arbor has two conditions, the text and the geometry, and each of them can be guided.
We denote the predicted velocity without conditions by $v_{\varnothing}$, with text by $v_y$, and with text and geometry by $v_{y,C}$.
We then sample with
\begin{equation}
\label{eq:guidance}
  \tilde{v} = v_{\varnothing} + s_t\,(v_y - v_{\varnothing}) + s_g\,(v_{y,C} - v_y).
\end{equation}
With $s_t = s_g$ this is standard classifier free guidance on the joint condition, and we use this setting for all our results.
Ctrl hardly changes between scales of $3$ and $10$ (Tab.~\ref{tab:supp:guidance}), so the results do not depend on a tuned value.
In Fig.~\ref{fig:supp:guidance} we show three constraints at four scales.
Without guidance some parts are lost, at slightly lower Ctrl.
With guidance the objects are complete and follow the same regions.

With separate scales a user could decide whether the prompt or the constraint takes precedence when the two disagree.

\begin{table}[t]
    \centering
    \small
    \setlength{\tabcolsep}{5pt}
    \begin{tabular*}{\columnwidth}{@{\extracolsep{\fill}}lccccc@{}}
        \toprule
        Guidance scale & none & 3 & 5 & \textbf{7.5} & 10 \\
        \midrule
        Ctrl, manual split    & 0.747 & 0.776 & 0.779 & 0.763 & 0.774 \\
        Ctrl, automatic split & 0.908 & 0.918 & 0.925 & 0.925 & 0.919 \\
        \bottomrule
    \end{tabular*}
    \caption{\textbf{Guidance scale.}
    Ctrl of Arbor for different guidance scales of the structure stage.
    The paper uses $7.5$.}
    \label{tab:supp:guidance}
\end{table}

\paragraph{Model.}
We attach Arbor to the sparse structure denoiser of TRELLIS-text-xlarge~\citep{trellis2024}.
We first voxelize the constraint meshes and encode them into sparse tokens with the frozen shape and material encoders of TRELLIS.2~\citep{trellis22025}.

A learned projection then maps the channels of these tokens to the width of the denoiser.
The router splits the latent grid into groups of neighboring queries, gives each group its nearest tokens, and appends the learned summary tokens.
Every block of the denoiser has one adapter.

\paragraph{Training.}
We train the geometry projection with its position embedding, the summary modules and the adapters.
During training we drop text and geometry independently, so that we can use classifier free guidance at sampling time.

\paragraph{Decoder adapters.}
The hull decoder is the frozen mesh decoder of TRELLIS with adapters at three of its levels.
These adapters read the same constraint tokens as the denoiser.
Here the router uses a region of one cell, so each decoder cell reads the tokens at its own position.
We train the adapters with the constraint sampler of Sec.~\ref{app:data_constraints} on structured latents that the SLAT stage sampled, so that they see the same kind of input in training and at inference.
Their losses act on geometry only, and we do not change the appearance branch of the decoder.

\subsection{Evaluation Details}\label{app:metrics}

\suppsummary{In this section we define all measures and describe the baselines and the user study. Our main measure Ctrl is the harmonic mean of hull overlap, avoidance, prompt retrieval and volume per object. A method therefore scores well only if it does well on all of them.}

\begin{table}[t]
    \centering
    \footnotesize
    \renewcommand{\arraystretch}{1.1}
    \setlength{\tabcolsep}{3pt}
    \newcommand{\sw}[1]{\textcolor{#1}{\rule{0.55em}{0.55em}}\,}
    \begin{tabular*}{\columnwidth}{@{\extracolsep{\fill}}l p{0.58\columnwidth} c@{}}
        \toprule
        Measure & What it asks & In Ctrl \\
        \midrule
        \sw{hull}Hull & How much of the hull surface does the output reach? & yes \\
        \sw{avoidance}Avoid & How deep does the output enter space that must stay empty? & yes \\
        PR & Can the output still be told apart as the object of its prompt? & yes \\
        Vol. & Does the output have as much solid volume as the source object, or does it fill the hull? & yes \\
        \midrule
        Avoid resp. & On which share of objects is no avoidance region entered? & no \\
        \sw{touch}Touch resp. & On which share of objects is every touch surface reached? & no \\
        F & How close is the output to the source object? & no \\
        Hull dev. & How far is the decoded surface from the hull at $256^3$? & no \\
        \bottomrule
    \end{tabular*}
    \caption{\textbf{The measures at a glance.}
    Ctrl is the harmonic mean of its four terms per object.
    Definitions are in Sec.~\ref{app:metrics}.}
    \label{tab:supp:terms}
\end{table}

\paragraph{Protocol.}
Both splits are fixed lists, and each entry holds the prompt, the constraint meshes and the seed.
Every method returns a mesh, and we voxelize and render all meshes with the same code in the cube $[-\tfrac12,\tfrac12]^3$.
Tab.~\ref{tab:supp:terms} lists all measures, and Fig.~\ref{fig:supp:metrics} illustrates the geometric ones on one object.

\begin{figure*}[!tbp]
    \centering
    \includegraphics[width=\textwidth]{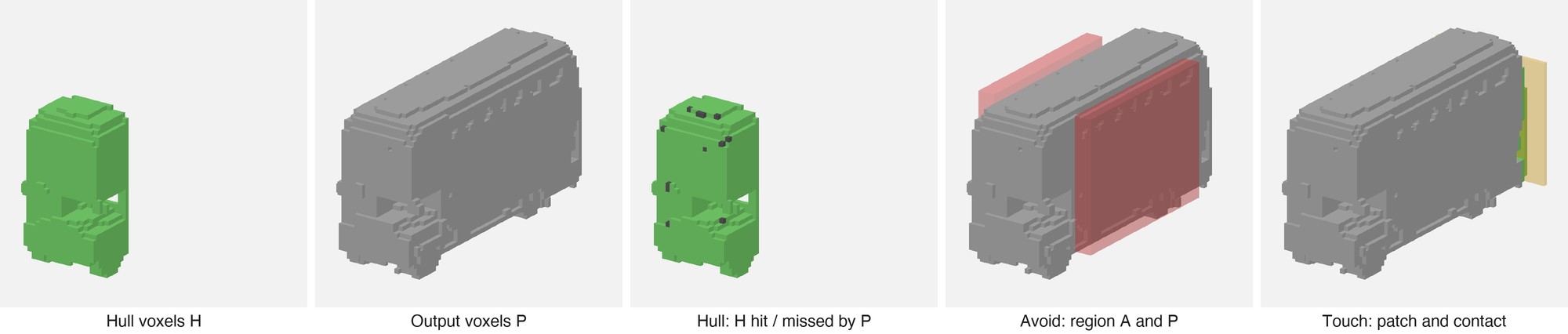}
    \caption{\textbf{The voxel sets behind Hull, Avoid and Touch.}
    The bus of Fig.~\ref{fig:comparison} with the output of Arbor on a $64^3$ grid.
    From left to right we show the hull voxels $H$, the output voxels $P$, the hull voxels that the output reaches and misses, the avoidance region $A$ and the touch patch.
    We use the $64^3$ grid for display, and Avoid itself is computed from surface points and a $256^3$ grid.}
    \label{fig:supp:metrics}
\end{figure*}

\paragraph{Hull.}
Hull measures how much of the hull the output reaches.
Let $H$ be the surface voxels of the hull and $P$ the surface voxels of the output on a $64^3$ grid.
We then compute the share of the hull voxels that the output reaches,
\[
\mathrm{Hull} = \frac{|P \cap H|}{|H|}.
\]

\paragraph{Avoid.}
Avoid measures how deep the output reaches into the regions that have to stay empty.
The avoidance region $A$ is the union of the avoidance meshes, voxelized at $256^3$.
We sample $200$k points on the output surface and compute the depth of each point inside $A$ with a Euclidean distance transform.
Let $q$ be the $99.9$th percentile of these depths, $\tau = 1/128$ the tolerance, which is one voxel at $128^3$, and $D$ the inscribed depth of $A$.
Avoid is then
\[
\mathrm{Avoid}
=
\begin{cases}
1, & q \le \tau,\\[2pt]
\max\!\Bigl(0,\ 1 - \dfrac{q-\tau}{D-\tau}\Bigr), & \text{otherwise}.
\end{cases}
\]
We say that an object \emph{respects} its avoidance regions if $\mathrm{Avoid}=1$.

\paragraph{Vol.}
A method could reach a high Hull by simply filling the whole cube.
Vol.\ guards against this by comparing the solid volume of the output with that of the source object.
For both we close the surface voxels at $64^3$ with one $3^3$ morphological closing and fill the interior.
With the resulting solids $V$,
\[
\mathrm{Vol}
=
\frac{\min\!\bigl(|V_{\text{pred}}|,\ |V_{\text{src}}|\bigr)}
{\max\!\bigl(|V_{\text{pred}}|,\ |V_{\text{src}}|\bigr)}.
\]
Note that Vol.\ is not a reconstruction metric, since it compares two voxel counts and never the shapes.
Before we scored any method, we fixed a test that Vol.\ itself had to pass.
The source latent decoded by TRELLIS had to reach a median of at least $0.85$ on both splits, and it reaches more than $0.99$.

\paragraph{Touch.}
We count a touch region as \emph{respected} if the output comes within $1/64$ of every touch patch and at most $0.1\%$ of its surface lies more than $1/64$ beyond the patch.

\paragraph{Prompt retrieval.}
A raw CLIP similarity separates the methods poorly, because gray renders of very different objects reach almost the same score.
With PR we therefore ask whether the renders of an output pick out their own prompt among all prompts of the same split.
We embed four gray views of the output with CLIP ViT-L/14~\citep{clip2021} and average the embeddings.
The rank score $\mathrm{AUC}_i$ is the fraction of the other prompts in the split whose text embedding scores below the prompt of the sample itself.
Chance gives $\mathrm{AUC}_i=0.5$, so we rescale it to
\[
\mathrm{PR}_i=\max\bigl(0,\ 2\,\mathrm{AUC}_i-1\bigr).
\]

\paragraph{Control score.}
We combine these terms into one score per object.
For object~$i$ the set of terms is $\mathcal{S}_i = \{\mathrm{Hull}_i,\ \mathrm{PR}_i,\ \mathrm{Vol}_i\}$, and objects with avoidance regions add $\mathrm{Avoid}_i$.
The object score is the harmonic mean
\[
\mathrm{Ctrl}_i
=
\frac{|\mathcal{S}_i|}{\sum_{s \in \mathcal{S}_i} 1/s},
\]
which is zero as soon as one term is zero.
A strong term therefore cannot make up for a failed constraint or for an output that lost its prompt.
All terms lie in $[0,1]$, and none of them is weighted or fitted.
In Sec.~\ref{app:more_numbers} we show that the ranking depends neither on the form of the avoidance term nor on the volume term.

\paragraph{Further measures.}
\emph{F} is the F-score between the output and the source object, without alignment.
\emph{Hull deviation} is the mean distance from $50$k points on the hull surface to the generated surface, in voxels of the $256^3$ grid.
As a measure of complexity we count the connected parts of a mesh that each cover at least one percent of its surface.
By this measure the meshes of Arbor are on par with those of TRELLIS (Tab.~\ref{tab:complexity}).

\begin{table}[t]
    \centering
    \footnotesize
    \setlength{\tabcolsep}{3pt}
    \newcommand{\gt}[1]{\textcolor{black!50}{#1}}
    \begin{tabular*}{\columnwidth}{@{\extracolsep{\fill}}lcc@{}}
        \toprule
        Method & Manual & Auto \\
        \midrule
        TRELLIS  & 1.91 & 2.10 \\
        \textbf{Arbor (ours)} & 1.97 & 1.91 \\
        \bottomrule
    \end{tabular*}
    \caption{\textbf{Mesh complexity.}
    Mean number of connected parts per mesh that each cover at least one percent of its surface.
    The interval of the difference between Arbor and TRELLIS includes zero on both splits.}
    \label{tab:complexity}
\end{table}

\paragraph{Baselines.}
TRELLIS is the released TRELLIS-text-xlarge model, which receives the prompt alone.
Gradient runs on the same model (Sec.~\ref{app:gradient_baseline}).
We run SpaceControl~\citep{spacecontrol2026} and Spice-E~\citep{spicee2024} with their official code and settings and give them the prompt and the merged hull mesh.
In Tab.~\ref{tab:hull_only} we repeat the comparison with Arbor restricted to this input.
SPAR3D and Hunyuan3D-Omni in Fig.~\ref{fig:recon_comparison} receive a render of the source object and points sampled from the hull.

\paragraph{User study.}
We ran the user study in a browser on the manual split.
Each trial shows the prompt, a render of the constraint, and four generated objects side by side, in random order and without method names.
The participants pick the object that best follows the prompt and the constraint.
In total $27$ participants answered $404$ trials.
Arbor appears in every trial and each baseline in a part of the trials.
The Pref.\ column of Tab.~\ref{tab:metrics} therefore gives, for each method, the share of the trials with this method in which its object was picked, and the column does not add up to $100$.
We also evaluate the choices in pairs, Arbor against each baseline, in Tab.~\ref{tab:user_study}.
Arbor is preferred over every baseline.
The participants gave their consent on the first page, before any trial, and confirmed that their anonymous answers and timings are stored for research.
Apart from two questions on their familiarity with 3D modeling and with generative AI, we collected no personal data.
The study was not reviewed by an ethics board.
We consider its risk minimal, since the participants only viewed and compared rendered 3D objects.

\begin{table}[t]
    \centering
    \small
    \setlength{\tabcolsep}{5pt}
    \begin{tabular*}{\columnwidth}{@{\extracolsep{\fill}}lc@{}}
        \toprule
        Arbor against & Share of choices for Arbor (\%) \\
        \midrule
        TRELLIS      & 59.7 [52.6, 66.4] \\
        Gradient     & 74.5 [67.4, 80.6] \\
        SpaceControl & 85.3 [78.8, 90.1] \\
        Spice-E      & 92.2 [86.8, 95.5] \\
        \bottomrule
    \end{tabular*}
    \caption{\textbf{User study by baseline.}
    Share of choices that went to Arbor among the trials in which either Arbor or the baseline was picked, with a $95\%$ Wilson interval.}
    \label{tab:user_study}
\end{table}

\subsection{Limitations}\label{app:limitations}

\paragraph{Resolution.}
Arbor can only control an object at the resolution of the generator it is attached to.
TRELLIS decides the structure on a $64^3$ grid and forms the surface at $256^3$, so detail below this grid cannot be controlled.
The decoder adapters recover part of this detail, and a generator with a finer grid would raise the limit.

\paragraph{Soft condition.}
The constraint is a learned condition, which means that it is not guaranteed to hold.
Most outputs follow it, but a sample can miss a part of the hull or touch an avoidance region.
In this case one can change the seed, since a new sample takes only seconds.
In Fig.~\ref{fig:supp:failures} we show three cases with a natural explanation.
The hull comes with normals that tell the model on which side the surface lies.
The model can still ignore them and take the hull as something inside the object, as for the two speaker hulls of the radio.
An avoidance region can also collide with the object of the prompt, so that no sensible object remains, and for the sink Arbor then generates only the taps and a drain.
An avoidance region very close to the hull can push the surface away from it, and for the hot plate only the rim stays on the hull.

\begin{figure*}[!tbp]
    \centering
    \includegraphics[width=\textwidth]{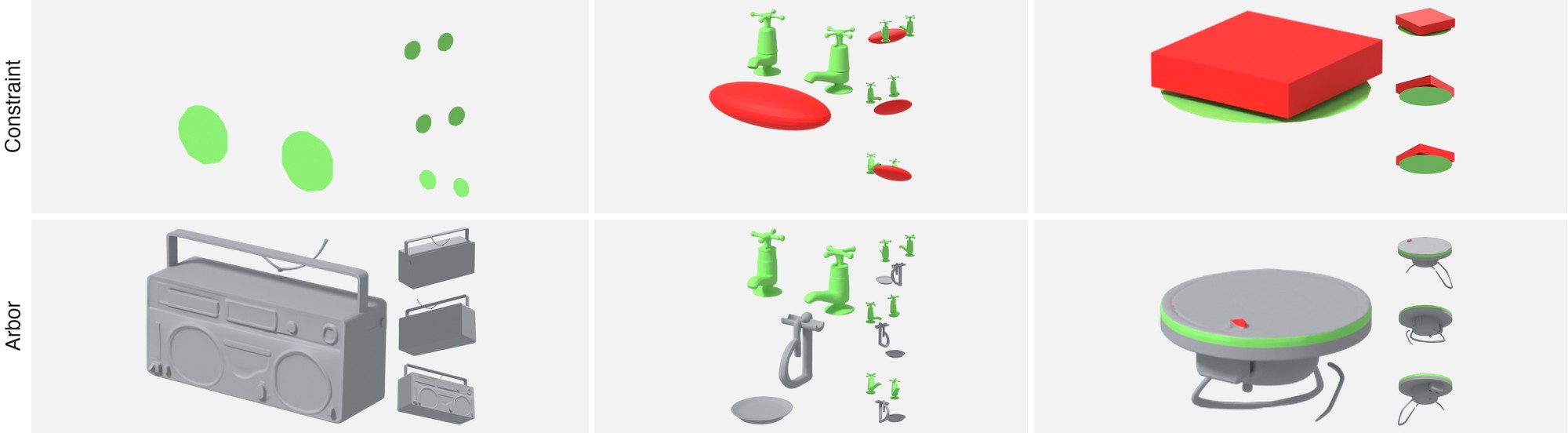}
    \caption{\textbf{Failure cases.}
    The speaker hulls of the radio are not part of the surface.
    The sink has only its taps and a drain.
    The hot plate follows the hull only at its rim.}
    \label{fig:supp:failures}
\end{figure*}

\paragraph{Types.}
We train Arbor with three types of regions.
Touch is the loosest of them, and it indicates that the recipe extends to further types, each of which needs its own training examples.
A natural next step is to attach semantic labels, so that a region can also state which part of the object it stands for.

\subsection{Questions and Answers}\label{app:qa}

In Tab.~\ref{tab:supp:qa} we give short answers to questions that a reader may have.

\begin{table*}[t]
    \centering
    \footnotesize
\renewcommand{\arraystretch}{1.12}
\setlength{\tabcolsep}{6pt}
\begin{tabular}{@{}p{0.25\textwidth}p{0.52\textwidth}p{0.16\textwidth}@{}}
    \toprule
    \textbf{Question} & \textbf{Short answer} & \textbf{Evidence} \\
    \midrule
    Why this architecture?
        & Guidance at sampling time needs no training but leaves rough and torn surfaces, and an additional loss on the constraint does not improve Ctrl. Arbor therefore learns the control inside the frozen generator with the plain training loss.
        & Sec.~\ref{app:variants} \\
    Why hull, avoidance and touch?
        & Hull and avoidance state where an object has to be and where it has to stay away. Touch is a looser form of control and shows that the same recipe carries over. A further type needs a new code and training examples.
        & Sec.~\ref{app:variants} \\
    Is the comparison fair if the baselines only receive a hull?
        & Yes. Arbor restricted to the same hull and prompt still scores above every baseline on both splits.
        & Tab.~\ref{tab:hull_only} \\
    Does the ranking depend on how Ctrl is defined?
        & No. Arbor ranks first under other forms of the avoidance term, without the volume term, without the prompt term, and for every combination of constraint types.
        & Tab.~\ref{tab:ctrl_sensitivity}, \ref{tab:ctrl_terms}, \ref{tab:ctrl_by_type} \\
    Does Ctrl reward copying the source object?
        & No. Ctrl has no reconstruction term, and Arbor also leads when the volume term, the only term that looks at the source object, is removed.
        & Tab.~\ref{tab:ctrl_terms} \\
    How certain is the ranking?
        & The paired difference between Arbor and every baseline lies above zero on both splits.
        & Tab.~\ref{tab:ctrl_ci} \\
    Does control survive at the decoded surface?
        & Yes. At $256^3$ Arbor reaches more of the hull than every baseline, and the hull decoder brings the surface onto the hull.
        & Tab.~\ref{tab:detail256_supp} \\
    Why a frozen pretrained encoder?
        & No encoder has to be trained, its tokens keep the detail of the constraint, and the same tokens also feed the decoder adapters.
        & Sec.~\ref{app:variants}, Tab.~\ref{tab:supp:detail256_enc60k} \\
    Why is there no loss on the constraint?
        & Each constraint is cut from its training object, so the flow matching loss already rewards reading it. Additional losses on the constraint did not improve Ctrl.
        & Sec.~\ref{app:variants} \\
    What is the router for?
        & It bounds what each block has to read. Its cap does not reduce control, and time and memory stay nearly flat as constraints grow.
        & Tab.~\ref{tab:router_k}, \ref{tab:router_scale} \\
    Is Arbor tied to TRELLIS?
        & Arbor needs tokens with positions, a router and an adapter in a transformer block. The same attachment already sits in two different networks, and all weights of TRELLIS stay frozen.
        & Sec.~\ref{app:variants} \\
    \bottomrule
\end{tabular}

    \caption{\textbf{Questions and answers.}}
    \label{tab:supp:qa}
\end{table*}

  \inputlocal{after_appendix.tex}
  {\small\bibliographystyle{ieeenat_fullname}\bibliography{references}}
\else
\twocolumn[{%
\renewcommand\twocolumn[1][]{#1}%
\maketitle
\vspace{-1.5em}
\begin{center}
    \centering
    \includegraphics[width=\textwidth]{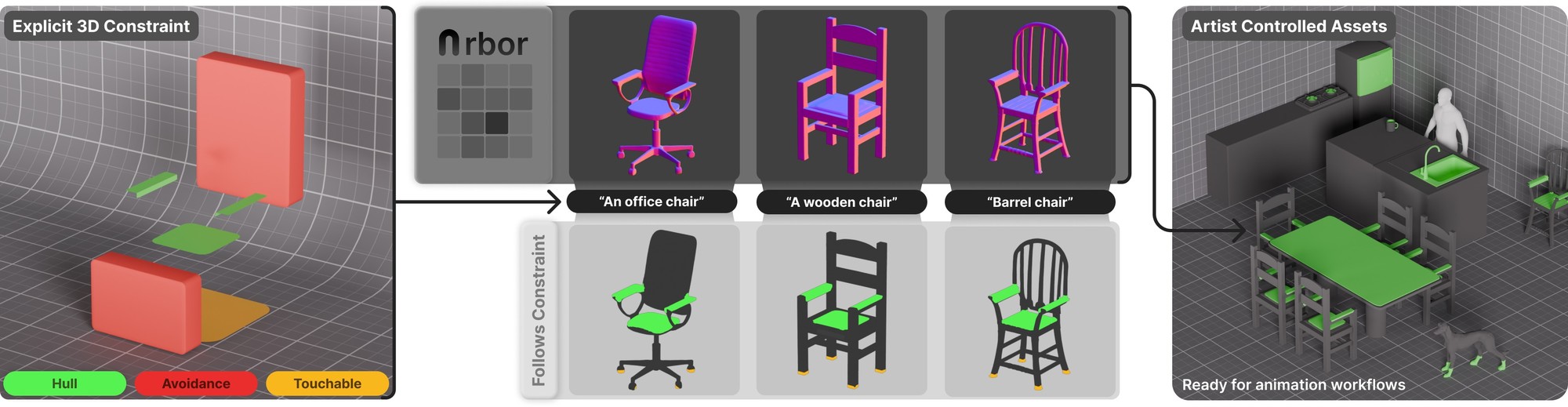}
    \captionof{figure}{\textbf{Arbor overview.}
    Arbor turns simple 3D control objects into an explicit constraint signal for text-conditioned 3D generation.
    Hull regions mark where generated geometry should exist, touch regions mark contact patches, and avoidance regions mark free space that should remain empty.
    This enables artist to co-author the generation process, making asset generation more reliable and therefore more likely to be used in production.
    Videos, code and model: \url{https://arbor.jdihlmann.com}.}
    \label{fig:introduction:overview}

\end{center}
\vspace{0.5em}
}]

\ifdefined\localdraft
  \input{content/workbench_legend}
\fi

\begin{abstract}
3D generators now turn a short prompt into a convincing asset, but they give the artist little control over the space an asset should occupy or avoid.
Yet this spatial intent is often known before generation starts.
An asset may have to fit a given place in a scene and leave room for what moves around it.
A prompt has no words for such a requirement, and an image fixes far more of the object than the artist wants to decide.
We present Arbor\footnote{Named after an arched support trellis to guide plant growth.}, a trained attachment that conditions a frozen 3D generator on authored constraint meshes.
Hull regions mark space the object has to occupy, and avoidance regions mark space it has to leave empty.
Arbor keeps these regions as geometry by converting the constraint meshes into tokens that the frozen generator reads.
To bound memory, a router hands each coarse 3D location only the part of the constraint that matters there.
The router plugs into the generator twice, first where the coarse structure of the object forms and then in the decoder that adds fine detail.
On automatic and hand authored benchmarks, Arbor follows the constraints far more closely than existing geometric control methods and preserves the quality of the generated meshes.

\end{abstract}

\inputlocal{after_abstract.tex}

\section{Introduction}\label{sec:introduction}

Latent 3D generators can now turn a short text prompt or a single image into a plausible asset~\citep{trellis2024,direct3d2024,hunyuan3d2025}.
For an artist who builds assets for a game, plausible is often not enough.
Consider a set of chairs that all share one sitting animation (Fig.~\ref{fig:introduction:overview}).
Every seat has to sit at the height and depth that the animation expects, the space under the seat has to stay free for the legs of the character, and every chair has to stand on the floor.
Everything else is open, from the style to the frame and the number of legs, and this open part is exactly where a generator should help.
With current generators the artist can only describe the chair, sample, and discard results until one happens to fit, then repair the rest by hand.
The requirement itself never reaches the generator.

The usual conditions do not carry such a requirement well.
A prompt names an object and its style, but it has no words for a seat at a given height or for a box of space that must stay empty.
An image does carry spatial information, yet the generator usually does not know the camera that produced it, so the depth and size of the object in 3D remain open.
The image also fixes the whole object and its appearance, far more than the few facts the artist actually knows, and every change of the requirement needs a new image.
Methods that condition generation on 3D geometry come closer.
Some guide the generator with a coarse stand in for the whole object, such as a skeleton, a few primitives or a rough shape~\citep{skadapter2026,coin3d2024,spicee2024,spacecontrol2026}.
The artist then has to block out the entire chair, although only the seat is known.
Others complete a partial shape~\citep{diffcomplete2023,pointsto3d2026,hunyuan3domni2025}.
The given part is an observation of the object that the generator completes.
None of these methods lets the artist mark space that has to remain empty, or define constraints of another kind (Tab.~\ref{tab:capabilities}).

\begin{table}[t]
    \centering
    \small
    \setlength{\tabcolsep}{0pt}
    \caption{\textbf{What the artist can specify.}
    Coarse shape: a rough stand in for the whole object. Parts only: geometry for some parts of the object, with the rest generated.}
    \label{tab:capabilities}
    \begin{tabular*}{\columnwidth}{@{\extracolsep{\fill}}lcccc@{}}
        \toprule
        Method & Text & \makecell{Coarse\\shape} & \makecell{Parts\\only} & \makecell{Empty\\space} \\
        \midrule
        \makecell[l]{SK-Adapter~\citep{skadapter2026}, Coin3D~\citep{coin3d2024},\\Spice-E~\citep{spicee2024}, SpaceControl~\citep{spacecontrol2026}} & \checkmark & \checkmark & -- & -- \\
        SPAR3D~\citep{spar3d2025} & -- & \checkmark & -- & -- \\
        Points-to-3D~\citep{pointsto3d2026} & \checkmark & -- & \checkmark & -- \\
        Hunyuan3D-Omni~\citep{hunyuan3domni2025} & -- & \checkmark & \checkmark & -- \\
        \midrule
        Arbor & \checkmark & \checkmark & \checkmark & \checkmark \\
        \bottomrule
    \end{tabular*}
\end{table}

Arbor adds this missing interface (Fig.~\ref{fig:introduction:overview}).
The artist states the requirement as meshes, prepared in a familiar modeling tool before generation.
A hull marks a volume the object has to fill, such as a seat, a handle or a wheel.
An avoidance region marks space that has to stay empty, such as a hole, a gap or the clearance for a moving part.
The same interface carries further kinds of constraints.
We show this with touch, which marks a surface the object has to reach, for example the ground under its legs.
The meshes can be plain blockout primitives or detailed parts, and the prompt and the generator decide everything they leave open.

Enforcing these constraints inside an existing generator is not straightforward.
The generator has to follow the meshes and still create the object the prompt names, and these two goals pull against each other.
If the constraint dominates, the output copies the meshes, and a plain box comes back where a sink cabinet was asked for (Fig.~\ref{fig:comparison}).
If the prior dominates, the constraint is ignored.
The constraint also has to enter the generator in the first place.
Modern generators operate on a compressed 3D latent~\citep{trellis2024,direct3d2024,hunyuan3d2025}, but the constraint meshes are detailed.
Two problems follow.
The meshes have to become compact tokens that keep local detail and the role of each region, and these tokens have to reach the latent regions where they matter.

Arbor solves the first problem with frozen encoders and the second with a trained router.
In summary, we make three contributions.
\begin{itemize}[leftmargin=1.25em, topsep=1pt, itemsep=1pt, parsep=0pt, partopsep=0pt]
\item \textbf{Encoded geometry as condition.} We show that the frozen surface encoders of TRELLIS.2~\citep{trellis22025} can be repurposed to turn constraint meshes and their roles into compact tokens that keep local structure and serve as model conditioning (Sec.~\ref{sec:method:encoding}).
\item \textbf{Geometry router and adapter.} The generator builds an object on a coarse 3D grid. Our router gives each region of this grid only the constraint tokens that lie near it, plus a short summary of the whole constraint, so memory stays bounded. An adapter then adds this information to the frozen generator. We attach both where the generator forms the coarse shape of the object and, optionally, in the decoder, so that the surface also follows the fine detail of a hull (Sec.~\ref{sec:method:routing}).
\item \textbf{Explicit geometric control interface.} The result is an interface in which the artist specifies hull and avoidance regions directly as 3D meshes, and which extends to further roles such as touch (Sec.~\ref{sec:method:interface}).
\end{itemize}

We instantiate Arbor on TRELLIS~\citep{trellis2024}, a text conditioned 3D generator whose weights stay frozen.
Training settles the balance between following a constraint and generating a convincing object.
No dataset exists that pairs objects with such constraints.
We therefore build our own training data by cutting each constraint out of a real object, and train Arbor to generate that whole object.
No new loss is needed for this, only the default training objective of the generator.
To assess constraint adherence and generation quality together, we introduce a control score.
Arbor reaches the highest score among all compared methods and respects nearly every avoidance region, and in a user study participants preferred its results most often (Sec.~\ref{sec:experiments}).

\inputlocal{after_introduction.tex}
\section{Related work}\label{sec:related}

\inlinesection{3D generation} has advanced along three practical lines.
Optimization based methods lift text or image supervision into 3D representations such as NeRFs or Gaussian splats~\citep{jain2022zeroshottextguidedobjectgeneration,poole2022dreamfusiontextto3dusing2d,lin2023magic3dhighresolutiontextto3dcontent,dreamcraft3d2023,tang2024dreamgaussiangenerativegaussiansplatting}.
Reconstruction based methods infer 3D from one or a few images and are often used for generation by first producing an image and then recovering 3D~\citep{lrm2023,triposr2024,instantmesh2024,sf3d2024,spar3d2025,dihlmann2026reli3drelightablemultiview3d}.
Native latent 3D models such as Direct3D~\citep{direct3d2024}, TRELLIS~\citep{trellis2024}, and Hunyuan3D~2.0~\citep{hunyuan3d2025} instead denoise compact 3D states directly.
Arbor instead brings the artist into the generation step itself, giving native latent generators the local spatial interface with roles that they currently lack.

ControlNet and Adapter methods showed that a pretrained text to image diffusion model can absorb new conditions through a control branch trained alongside the frozen backbone~\citep{controlnet2023,mou2023t2iadapterlearningadaptersdig,ye2023ipadaptertextcompatibleimage}.
The 3D version of the problem is harder, because a condition must remain meaningful across viewpoints, compressed latent states, and the final output representation.
Arbor inherits the attached control intuition, but rebuilds it for native 3D generation where the condition is itself geometry rather than a 2D map.

\inlinesection{Editing}
methods receive an existing asset or scene up front and modify it in a chosen region.
SIGNeRF edits NeRF scenes with depth conditioned reference sheets~\citep{signerf2024}.
Instant3dit and ObjFiller3D inpaint 3D objects through multiview image edits before mapping the result back to 3D~\citep{instant3dit2025,objfiller3d2025}, while Easy3E performs feed forward asset editing in the sparse voxel latent of TRELLIS~\citep{easy3e2026}.
Each method requires the user to bring the asset they want to modify.
Arbor instead conditions a generator before any asset exists, with only typed local constraint regions on top of a text prompt.

\inlinesection{Reconstruction}
methods condition on an input that already describes the asset's global structure.
SK-Adapter is closest in architecture, since it also trains cross attention with a zero-initialized residual update inside the frozen sparse-structure transformer of TRELLIS, but its tokens encode one skeleton (joint positions and topology), with no roles for regions and no input for space that has to stay empty~\citep{skadapter2026}.
Coin3D conditions on a coarse primitive proxy through a 3D adapter~\citep{coin3d2024}.
Others use a retrieved or user provided 3D reference shape~\citep{phidias2025}.
SPAR3D conditions on an image plus an editable point cloud that it first predicts from that image~\citep{spar3d2025}.
Hunyuan3D-Omni combines an image with one of several geometric modalities, namely points, boxes, voxels, or skeletons.
It accepts partial observations but is built around an image condition~\citep{hunyuan3domni2025}.
Each input describes the asset's global structure ahead of generation, whether as a skeleton, a proxy, a reference, or an image.
Arbor specifies only the parts that matter and leaves the rest to the prompt and the generator, allowing different assets to satisfy the same constraint set.

\inlinesection{Inpainting and completion}
methods start from sparse, partial or coarse geometry and grow it into a finished asset.
DiffComplete and Points-to-3D condition a 3D diffusion model on partial geometry and complete the rest around the input surface~\citep{diffcomplete2023,pointsto3d2026}, while Spice-E uses cross entity attention between a guidance shape, such as a coarse primitive abstraction, and the noisy sample~\citep{spicee2024}.
SpaceControl steers a pretrained 3D denoiser during sampling without extra training, exposing a global tradeoff between constraint fidelity and realism~\citep{spacecontrol2026}.
All of them treat the input geometry as a single signal to complete, match, or steer toward.
Arbor instead gives each region its own role.
Hull regions indicate where the asset should exist, touch regions mark contact surfaces, and avoidance regions are defined precisely by not becoming surface.
The artist authors these regions as ordinary meshes, from rough blockout boxes to detailed parts, and constrains only the parts of the asset that matter.
Arbor follows them in the coarse structure of the asset and, through its decoder, down to the fine detail of a hull.

\inputlocal{after_related_work.tex}
\section{Method}\label{sec:method}

The main goal of Arbor is the creation of explicit geometry control methods for 3D generation. Given a text prompt~$y$ and a constraint object~$C$, the goal is to sample a 3D asset~$x \sim p(x \mid y, C)$.
Fig.~\ref{fig:pipeline} follows the constraint through the model, and the circled numbers refer to its steps.
The hull, avoidance and touch meshes of the artist are first fused into one constraint object~\step{0}.
A frozen encoder turns it into geometry tokens that know their position and their type~\step{1}\,\step{2}.
A router then gives each region of the generator's latent grid the tokens that lie near it~\step{3}, and an adapter in every block of the frozen generator reads them and adjusts the latent state~\step{4}.
Only the router and the adapters are trained.

\begin{figure*}[t]
    \centering
    \includegraphics[width=\textwidth]{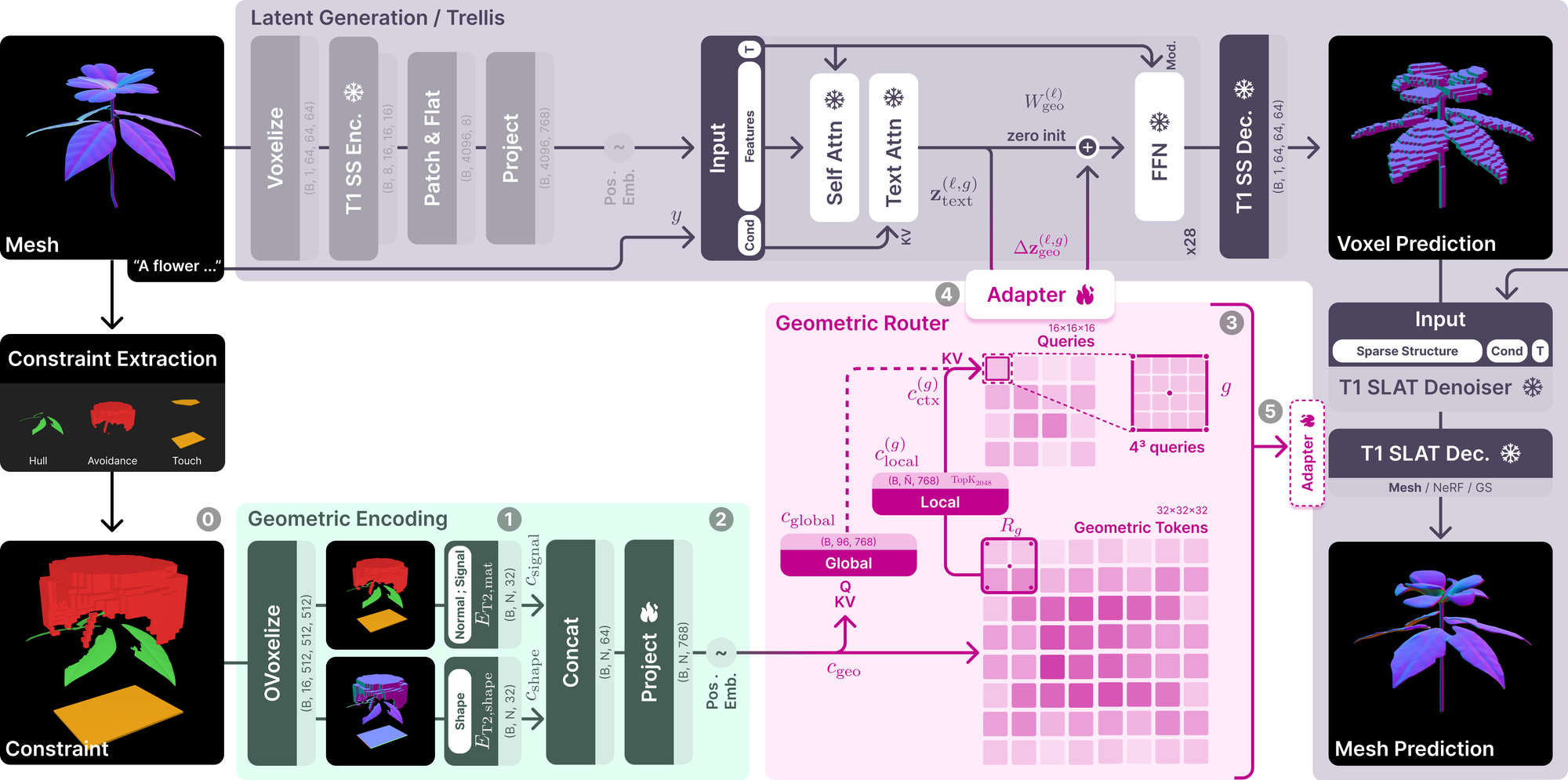}
    \caption{\textbf{Constraint conditioning pipeline.}
    The constraint shown is cut automatically from the mesh, as in our training data.
    \protect\step{0}~Hull, avoidance and touch meshes are fused into one constraint object.
    \protect\step{1}~The frozen T2 encoders turn it into shape tokens $c_{\mathrm{shape}}$ and signal tokens $c_{\mathrm{signal}}$.
    \protect\step{2}~A learned projection and positional embedding give the geometry tokens $c_{\mathrm{geo}}$.
    \protect\step{3}~The router gives each of the $64$ groups of T1 queries its $2048$ nearest tokens $c_{\mathrm{local}}^{(g)}$ and $96$ global summary tokens $c_{\mathrm{global}}$.
    \protect\step{4}~In every frozen T1 block, the adapter adds a zero initialized geometry update $\Delta\mathbf{z}_{\mathrm{geo}}^{(\ell,g)}$ before the FFN.
    \protect\step{5}~The same attachment can be added to the T1 mesh decoder.}
    \label{fig:pipeline}

\end{figure*}

\subsection{Background}

Arbor needs a generator that is driven by text, so that the prompt can decide everything the constraint leaves open.
Many recent latent 3D generators, including TRELLIS.2~\citep{trellis22025} and Hunyuan3D~2.0~\citep{hunyuan3d2025}, are conditioned on an image only.
We therefore build on TRELLIS~\citep{trellis2024}, a strong open text conditioned generator, which we call T1 and use in its xlarge variant (top row of Fig.~\ref{fig:pipeline}, Latent Generation).
We refer to the original paper for details and summarize only what Arbor needs.
T1 generates in two stages, a sparse structure (SS) stage followed by a structured latent (SLAT) refinement stage.
The SS stage denoises a $16^3$ latent grid, which is decoded into a $64^3$ occupancy grid and thereby decides the coarse structure of the object. This is why Arbor attaches to it.
In simplified form, each of its transformer blocks updates the latent state by self attention, text cross attention and a feed forward network,
\begin{equation}
\label{eq:t1-block}
\begin{gathered}
  z_{\mathrm{text}}^{(\ell)} = \mathrm{CrossAttn}_{\mathrm{text}}^{(\ell)}\!\Bigl(\mathrm{SelfAttn}^{(\ell)}(z^{(\ell-1)}), y\Bigr), \\
  z^{(\ell)} = \mathrm{FFN}^{(\ell)}\!\bigl(z_{\mathrm{text}}^{(\ell)}\bigr),
\end{gathered}
\end{equation}
where $\ell$ indexes the block and $y$ is the encoded text prompt.

\subsection{Constraints}
\label{sec:constraints}\label{sec:method:interface}

Constraints in Arbor are 3D meshes that specify desired spatial behavior.
Hull and avoidance are our main constraints, and touch shows that the same recipe extends to further kinds of control.

\paragraph{Types}
\label{sec:constraints:types}
Hull constraints define regions where generated geometry should exist, e.g., the seat of a chair.
Avoidance constraints define regions where generated geometry should not exist, e.g., empty space above the seat.
Touch constraints define surfaces the object should contact, e.g., chair legs touching the ground.
Each type can cover any number of separate regions, from coarse blockout primitives to detailed meshes, in the normalized object frame of T1 (Appendix~\extref{app:data_constraints}).

\subsection{Constraint Encoding}
\label{sec:constraints:encoding}\label{sec:method:encoding}

The encoder of the constraint has to read a mesh directly, keep its local detail, and offer a place for the constraint types.
The natural first option would be to encode constraints with the native T1 encoders, since T1 is the generator we want to influence.
However, its structure encoder reads only a $64^3$ occupancy grid, and its structured latent encoder needs a complete object that is voxelized, rendered from $150$ views, and projected with DINOv2~\citep{dinov2} features, which is slow and leaves no direct place for typed control signals.
We also tried to train our own encoder, but found that a pretrained geometric encoder gives more usable control.
The reason is a chicken and egg problem. Our adapters learn to pick up what the tokens offer, but an untrained encoder offers nothing to pick up. Only a reconstruction loss on the constraint could train it, which is not a good choice (Sec.~\ref{sec:method:training}).

We therefore settle for the frozen encoders of TRELLIS.2 (T2)~\citep{trellis22025}.
T2 first converts a mesh into OVoxels, a sparse voxel representation that stores geometry together with aligned surface attributes.
Its encoders then compress the OVoxels into sparse tokens.
The attribute channels give us the place for the constraint types.
We are aware that using two TRELLIS models can be confusing at first.
To be clear, T1 remains the only generator in Arbor.
From T2 we take nothing but its frozen encoders.

To get more specific, Arbor fuses the separate constraint meshes into a single mesh $C_{\mathrm{mesh}}$ with surface normals $C_{\mathrm{normals}}$ and signal channels $C_{\mathrm{signal}}$ that encode the constraint type~\step{0}.
OVoxelizing this constraint object yields the shape field $\widetilde{C}_{\mathrm{shape}}$ and aligned fields $\widetilde{C}_{\mathrm{normals}}$ and $\widetilde{C}_{\mathrm{signal}}$ on a sparse $512^3$ voxel grid.
The shape encoder receives $\widetilde{C}_{\mathrm{shape}}$, while the attribute encoder is repurposed~\step{1}.
Instead of the original $6$ material channels, it receives $3$ voxel aligned normal channels $\widetilde{C}_{\mathrm{normals}}$ and $3$ binary control channels for typed signals $\widetilde{C}_{\mathrm{signal}}$.
The normals tell the encoder which side of a thin region is meant, for example the top of a seat rather than its underside.
The three flags mark the type of every surface point and do not limit how many regions a constraint has.
\begin{equation}
\label{eq:constraint-encoding}
\begin{gathered}
  c_{\mathrm{shape}} = E_{\mathrm{T2,shape}}\!\bigl(\boldsymbol{\widetilde{C}_{\mathrm{shape}}}\bigr), \\
  c_{\mathrm{signal}} = E_{\mathrm{T2,mat}}\!\bigl( \boldsymbol{\widetilde{C}_{\mathrm{normals}}} \,;\, \boldsymbol{\widetilde{C}_{\mathrm{signal}}} \bigr).
\end{gathered}
\end{equation}
Both encoders reduce the resolution by a factor of $16$, so $c_{\mathrm{shape}}$ and $c_{\mathrm{signal}}$ are sparse $32$-dimensional tokens on a $32^3$ grid.
We concatenate both latent streams into the geometry memory $c_{\mathrm{geo}}=\bigl(c_{\mathrm{shape}} \; c_{\mathrm{signal}}\bigr)$.
Each of the $N$ geometry tokens keeps its OVoxel position $p_i$, and $N$ grows with the surface area of the constraint.
While this encoding is simple and effective, two issues remain.
The tokens are not native to the T1 latent space, and they live on a finer $32^3$ grid than the T1 $16^3$ state.
We map every geometry token to the T1 model width with a learned projection and add a learned 3D positional embedding from $p_i$~\step{2}.
Arbor never converts these tokens into T1 latents. They stay an external memory that T1 reads through attention.
The router handles the resolution mismatch by choosing the tokens each query group receives.

\subsection{Control}
\label{sec:constraints:control}\label{sec:method:routing}\label{sec:method:injection}

Generated geometry should obey the constraint, but still follow the text prompt.
These goals can conflict.
A model can improve hull overlap by overfilling the guide while losing prompt fit, structure, or variation.
This is why sampling time steering~\citep{spacecontrol2026}, including our Gradient baseline (Appendix~\extref{app:gradient_baseline}), are an important but incomplete alternative.
Arbor instead learns the attachment inside the generator.
Each SS block carries one hidden query token for every cell of the $16^3$ lattice.
Arbor partitions this query lattice into $64$ groups indexed by $g$, each covering a $4 \times 4 \times 4$ block of neighboring queries.
With cells indexed by $(i,j,k) \in \{0,\ldots,15\}^3$, group $g \in \{0,\ldots,3\}^3$ is
\begin{equation}
\label{eq:groups}
  G_g = \bigl\{(i,j,k) \;\big|\; \bigl(\lfloor i/4 \rfloor, \lfloor j/4 \rfloor, \lfloor k/4 \rfloor\bigr) = g \bigr\}.
\end{equation}
The groups are disjoint, so every query belongs to exactly one group and keeps its position in the lattice.

\paragraph{Router}
The main challenge is scale, since Arbor conditions on dense encoded geometry rather than the small token set used by skeleton adapters such as SK-Adapter~\citep{skadapter2026}.
Attending from all T1 queries to all constraint tokens would be expensive and would require truncation.
The router makes dense geometry memory usable by the SS denoiser~\step{3}.
At block $\ell$, the original T1 stream supplies the attention queries, namely the text conditioned hidden states $\mathbf{z}_{\mathrm{text}}^{(\ell,g)}$ of group $g$.
The router supplies the keys and values by retrieving local geometry for each query group and appending a compact global context.
All queries in one group share the context,
\begin{equation}
\label{eq:router-context}
  c_{\mathrm{ctx}}^{(g)} = \bigl[c_{\mathrm{local}}^{(g)} \,;\, c_{\mathrm{global}}\bigr].
\end{equation}
For local routing, each group uses its center and eight corners as routing anchors $R_g$.
We score every geometry token by its distance to the nearest anchor and keep the $K=2048$ nearest tokens,
\begin{equation}
\label{eq:router-local}
  c_{\mathrm{local}}^{(g)} = \operatorname{TopK}_{2048}\Bigl(\{(c_{\mathrm{geo}}^{(i)}, p_i)\}_{i=1}^{N}, \min_{r \in R_g} \lVert p_i - r \rVert_2\Bigr).
\end{equation}
This TopK step is not attention.
It selects a bounded local memory for the adapter, keeping cost fixed while tying the context to the relevant constraint region.
To retain object level information that local routing can miss, Arbor also summarizes the full geometry memory with $96$ learned summary tokens,
\begin{equation}
\label{eq:router-global}
  c_{\mathrm{global}} = \operatorname{MLP}\Bigl(\operatorname{CrossAttn}\bigl(S_{\mathrm{global}}, \{c_{\mathrm{geo}}^{(i)}\}_{i=1}^{N}\bigr)\Bigr),
\end{equation}
where $S_{\mathrm{global}}$ is the learned summary set.
Each group therefore reads at most $2048$ local and $96$ global tokens, so the attention in every block stays bounded however large the constraint is.
Only the global summary reads all tokens, once per constraint, so the total cost grows linearly and mildly with the size of the constraint.
Control is insensitive to $K$ over a wide range, so this cap costs no control, and the structure stage takes about six seconds and under $10$\,GB of memory per object (Appendix~\extref{app:router_k}).

\begin{figure*}[t]
    \centering
    \includegraphics[width=\textwidth]{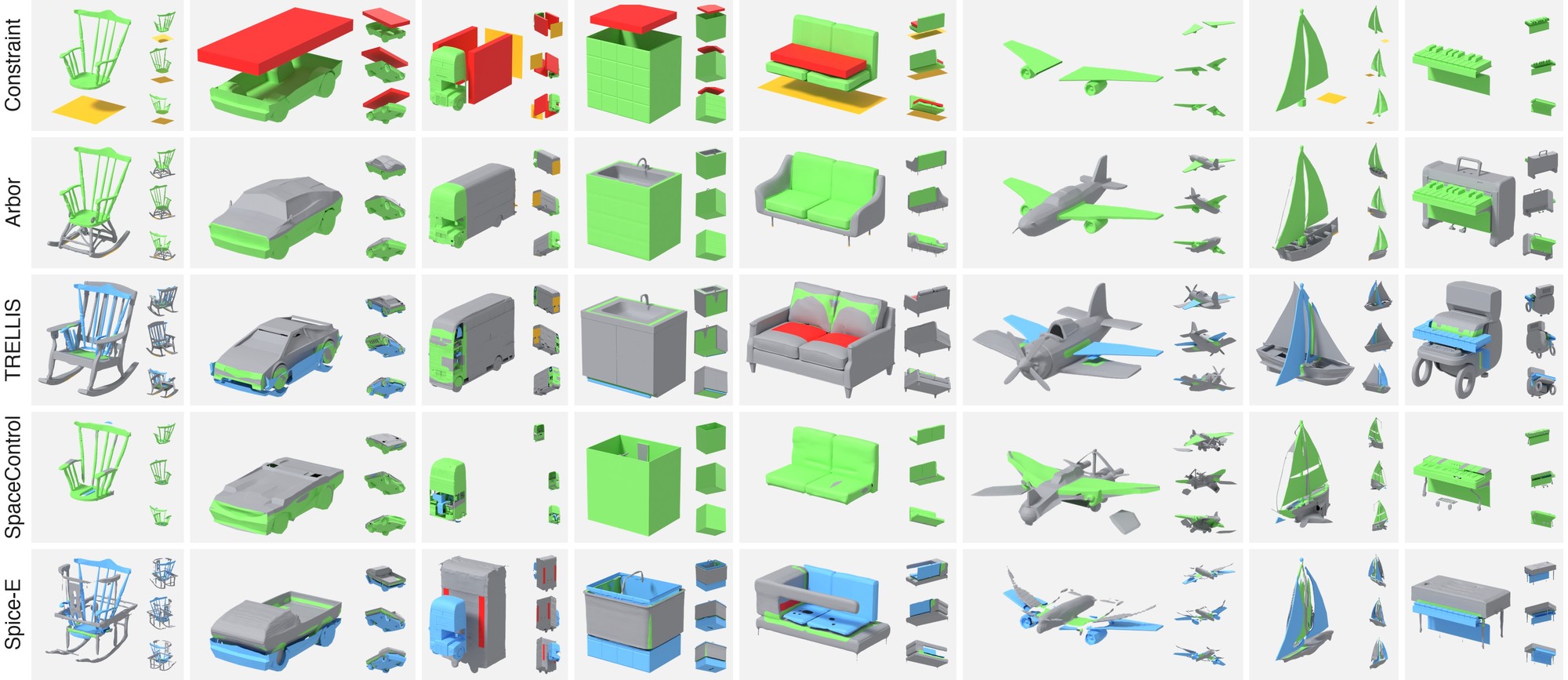}
    \caption{\textbf{Controlled generation compared with geometric control baselines.}
    Each column is one constraint and prompt; all methods use the same constraint, prompt and seed.
    The top row shows the constraint geometry with its regions coloured
    \textcolor[rgb]{0.34,0.85,0.32}{hull}, \textcolor[rgb]{0.9,0.18,0.18}{avoidance} and \textcolor[rgb]{0.95,0.7,0.1}{touch}.
    In every method row, the generated surface is coloured where it lies on or in a region: green on the hull, red inside avoidance, yellow on touch,
    and \textcolor[rgb]{0.35,0.7,1.0}{blue} marks hull that the output leaves empty.}
    \label{fig:comparison}

\end{figure*}

\begin{table*}[t]
    \centering
    \small
    \renewcommand{\arraystretch}{1.0}
    \setlength{\tabcolsep}{4.2pt}
    \newcommand{\best}[1]{\cellcolor{cvprblue!14}\textbf{#1}}
    \newcommand{\second}[1]{\underline{#1}}
    \newcommand{\gt}[1]{\textcolor{black!50}{#1}}
    \newcommand{\sw}[1]{\textcolor{#1}{\rule{0.55em}{0.55em}}\,}
    \newcommand{\hd}[1]{\makecell[c]{#1}}
    \resizebox{\linewidth}{!}{%
    \begin{tabular}{@{}l c @{\hspace{1.0em}} c cccc cc @{\hspace{1.0em}} c cccc cc@{}}
    \toprule
    & & \multicolumn{7}{c}{\textbf{Manual split}} & \multicolumn{7}{c}{\textbf{Automatic split}} \\
    \cmidrule(lr){3-9} \cmidrule(l){10-16}
    & & & \multicolumn{4}{c}{\scriptsize terms of Ctrl} & \multicolumn{2}{c}{\scriptsize not in Ctrl}
        & & \multicolumn{4}{c}{\scriptsize terms of Ctrl} & \multicolumn{2}{c}{\scriptsize not in Ctrl} \\[-0.9ex]
    \cmidrule(lr){4-7} \cmidrule(lr){8-9} \cmidrule(lr){11-14} \cmidrule(l){15-16}
    Method & Pref.\,\% & \textbf{Ctrl} & \sw{hull}Hull & \sw{avoidance}Avoid & PR & Vol. & \hd{\sw{avoidance}Avoid\\resp.} & \hd{\sw{touch}Touch\\resp.}
           & \textbf{Ctrl} & \sw{hull}Hull & \sw{avoidance}Avoid & PR & Vol. & \hd{\sw{avoidance}Avoid\\resp.} & \hd{\sw{touch}Touch\\resp.} \\
    \midrule
    TRELLIS        & \second{33.3} & 0.403 & 0.283 & 0.748 & \best{0.946} & \second{0.599} & 0.667 & \second{0.714}
                   & 0.436 & 0.269 & 0.704 & \best{0.953} & \second{0.688} & 0.266 & 0.333 \\
    Gradient       & 18.2 & \second{0.596} & 0.690 & 0.798 & 0.908 & 0.563 & 0.500 & 0.571
                   & \second{0.697} & \second{0.775} & 0.687 & 0.867 & 0.645 & 0.281 & \second{0.571} \\
    SpaceControl   & 9.6 & 0.487 & \second{0.823} & \second{0.962} & 0.782 & 0.362 & \second{0.917} & 0.000
                   & 0.631 & 0.729 & \second{0.890} & 0.855 & 0.581 & \second{0.641} & 0.333 \\
    Spice-E        & 5.2 & 0.202 & 0.108 & 0.635 & 0.730 & 0.594 & 0.417 & 0.286
                   & 0.321 & 0.196 & 0.676 & 0.795 & 0.615 & 0.312 & 0.381 \\
    \midrule
    \textbf{Arbor (ours)} & \best{59.2} & \best{0.763} & \best{0.899} & \best{1.000} & \second{0.933} & \best{0.700} & \best{1.000} & \best{0.857}
                   & \best{0.925} & \best{0.950} & \best{0.999} & \second{0.946} & \best{0.904} & \best{0.969} & \best{0.921} \\
    \midrule
    \gt{GT latent, decoded} & \gt{--} & \gt{0.895} & \gt{0.918} & \gt{0.972} & \gt{0.964} & \gt{0.928} & \gt{0.917} & \gt{1.000}
                   & \gt{0.958} & \gt{0.941} & \gt{1.000} & \gt{0.955} & \gt{0.986} & \gt{1.000} & \gt{1.000} \\
    \bottomrule
    \end{tabular}%
    }
    \caption{\textbf{Controlled generation on Toys4K} (32 hand-authored and 128 automatic layouts).
    Ctrl is the per object harmonic mean of its terms; \emph{resp.}\ is the share of objects whose regions are fully respected.
    Pref.\ is the win rate in a user study (27 participants, 404 trials), pooled over the rated Arbor variants.
    Gray: decoded ground-truth latents. Definitions in Appendix~\extref{app:metrics}; best \textbf{bold} on blue, second \underline{underlined}.}
    \label{tab:metrics}
\end{table*}

\paragraph{Adapter}
Arbor injects geometry with a separate grounding branch inside each T1 block, after frozen text cross attention and before the feed forward (FFN) update~\step{4}.
Eq.~\ref{eq:adapter} modifies the T1 block from Eq.~\ref{eq:t1-block} by inserting a geometry residual before the original FFN.
For the queries in group $g$ at block $\ell$, with router context $c_{\mathrm{ctx}}^{(g)}$ from Eq.~\ref{eq:router-context}, we write
\begin{equation}
\label{eq:adapter}
\begin{gathered}
  \Delta \mathbf{z}_{\mathrm{geo}}^{(\ell,g)} = \mathrm{CrossAttn}_{\mathrm{geo}}^{(\ell)}\!\Bigl(\mathbf{z}_{\mathrm{text}}^{(\ell,g)}, c_{\mathrm{ctx}}^{(g)}\Bigr), \\
  \mathbf{z}^{(\ell,g)} = \mathrm{FFN}^{(\ell)}\!\Bigl(\mathbf{z}_{\mathrm{text}}^{(\ell,g)} + W_{\mathrm{geo}}^{(\ell)} \Delta \mathbf{z}_{\mathrm{geo}}^{(\ell,g)}\Bigr).
\end{gathered}
\end{equation}
Here $\mathbf{z}_{\mathrm{text}}^{(\ell,g)}$ denotes the subset of text conditioned hidden states in query group $g$, $\Delta \mathbf{z}_{\mathrm{geo}}^{(\ell,g)}$ is the geometry update, and $W_{\mathrm{geo}}^{(\ell)}$ is zero initialized.
Queries and update both have shape $64 \times 1280$, where $1280$ is the width of T1, and the update is written back to the cells of the group, so nothing is pooled or merged across groups.
$\mathrm{CrossAttn}_{\mathrm{geo}}^{(\ell)}$ is one packed attention pass over local tokens and global summary tokens.
This placement keeps the pretrained text path intact while allowing geometry to affect the state before the FFN.
Because $W_{\mathrm{geo}}^{(\ell)}$ starts at zero and all T1 weights stay frozen, Arbor without a constraint is exactly T1.

\paragraph{Training.}\label{sec:method:training}
We train only the geometry facing modules.
These are the geometry projection and position embedding, the global summary modules, the semantic part token modules, and the grounding adapters.
Optimization uses the standard SS flow matching objective of the T1 backbone.
We tried adding a reconstruction loss on the constraint, but it trades the prior for the constraint, and the model then reconstructs the constraint rather than a full object.
Text and geometry conditions are dropped independently for classifier free guidance~\citep{cfg2022}.
Each training example uses the original T1 sparse structure latent of an object as the flow matching target.
No dataset pairs objects with such constraints, so we sample them online from about $50$k objects of ABO~\citep{abo2022}, HSSD~\citep{hssd2024} and the Sketchfab subset of Objaverse-XL~\citep{objaversexl2023}.
Every constraint is cut from the object that serves as the target, so it is satisfiable by construction.
Details, schedules, and implementation settings are deferred to Appendix~\extref{app:data_constraints}.
At inference we use classifier free guidance on the joint text and geometry condition (Eq.~\extref{eq:guidance} in the supplement).

\paragraph{Decoder.}\label{sec:method:decoder}
The SS stage fixes where the object is at $64^3$, and the SLAT stage of T1 runs unchanged on this structure.
The mesh decoder then forms the surface at $256^3$, where thin hull detail can still be lost.
We therefore place the same attachment inside the frozen mesh decoder~\step{5}.
Zero initialized adapters read the routed constraint tokens and are trained with geometry losses alone.
This second attachment is optional, recovers fine hull detail at $256^3$ and leaves the $64^3$ structure unchanged (Sec.~\ref{sec:experiments:detail}).

\inputlocal{after_method.tex}

\begin{figure*}[t]
    \centering
    \includegraphics[width=\textwidth]{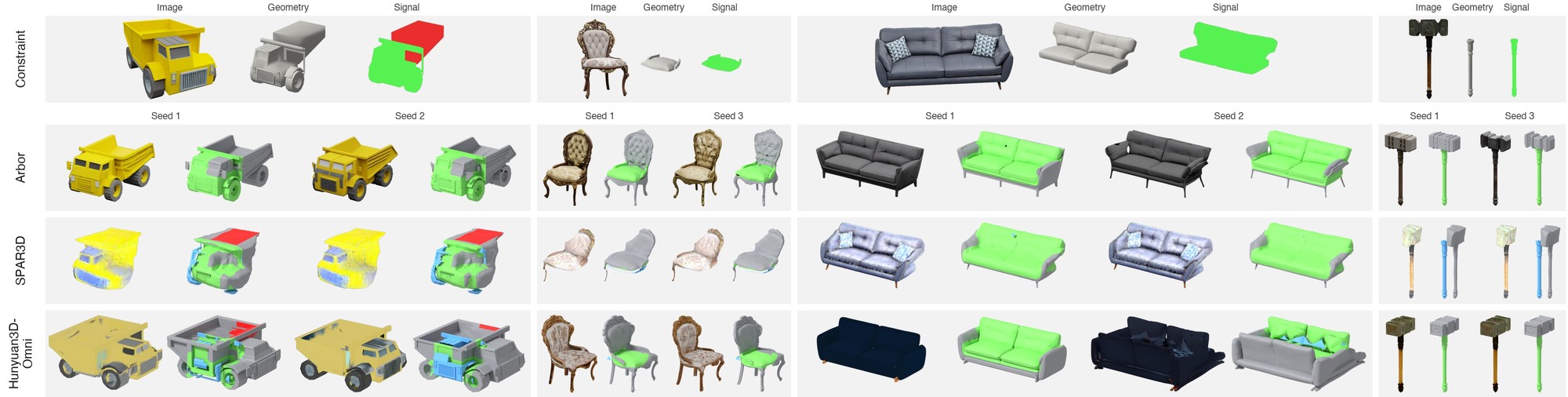}
    \caption{\textbf{Comparison with image-conditioned reconstruction methods.}
    Top: source object, constraint geometry and its regions
    (\textcolor[rgb]{0.34,0.85,0.32}{hull}, \textcolor[rgb]{0.9,0.18,0.18}{avoidance}, \textcolor[rgb]{0.95,0.7,0.1}{touch}).
    SPAR3D and Hunyuan3D-Omni receive a render of the object and points sampled from the hull, but not the avoidance or touch regions; they solve reconstruction from an image and are shown for contrast, not as failed control methods.
    Each method shows two seeds, the textured result next to its constraint following
    (green on the hull, red inside avoidance, yellow on touch, \textcolor[rgb]{0.35,0.7,1.0}{blue} for hull left empty).
    Arbor varies across seeds while filling the hull and staying out of avoidance; the baselines stay close to their image and enter the avoidance region of the truck.}
    \label{fig:recon_comparison}
\end{figure*}

\section{Evaluation}\label{sec:experiments}

We evaluate whether Arbor follows authored constraints while still producing the object the prompt asks for.
A method that copies the hull into the output passes the first test and fails the second, and a text generator that ignores the constraint does the opposite.
Our metrics therefore score both at once, on every object.

\textbf{Models and baselines.}
Arbor is the attachment of Sec.~\ref{sec:method} on TRELLIS-text-xlarge~\citep{trellis2024}, including the decoder adapters.
TRELLIS is this generator without constraints.
SpaceControl~\citep{spacecontrol2026} and Spice-E~\citep{spicee2024} are official external models.
No existing method accepts forbidden or contact regions, so they receive the merged hull, the closest input they support.
Gradient is our baseline for guidance at sampling time on the same frozen generator, without any training. During the last denoising steps it decodes the structure and follows the gradient of a hull occupancy loss (Appendix~\extref{app:gradient_baseline}).

\textbf{Benchmarks.}
We evaluate on Toys4K~\citep{toys4k2021}, which TRELLIS also holds out for evaluation~\citep{trellis2024} and Arbor never sees in training.
The large public 3D corpora are part of the training data of the backbone and would not give a clean test.
The automatic split pairs $128$ objects with constraints drawn by the training sampler.
The manual split contains $32$ hand authored constraint sets that are not drawn by the training sampler, such as blockout boxes, rough frames and forbidden slabs (Appendix~\extref{app:data_constraints}).
Each set is modelled by hand in Blender for one object, with the care an artist would take, so this split is small by design.

\textbf{Metrics.}
Hull is the share of the hull surface that the output reaches on a $64^3$ grid.
Avoid scores how deep the output enters an avoidance region. It is one when almost no surface lies more than one voxel at $128^3$ inside the region, and falls to zero at the full depth of the region.
PR asks whether gray renders of the output pick out their own prompt among all prompts of the split, rescaled so that chance scores zero.
Vol.\ compares the solid volume of the output with that of the source object and guards against outputs that simply fill the hull.
It is a soft term, since several shapes can satisfy the same constraint.
Ctrl is the harmonic mean of these terms per object, so an object scores well only if it does well on all of them, and we report its mean over the split.
Next to Ctrl we report how many objects respect their avoidance and touch regions completely.

\begin{figure}[!tbp]
    \centering
    \includegraphics[width=\columnwidth]{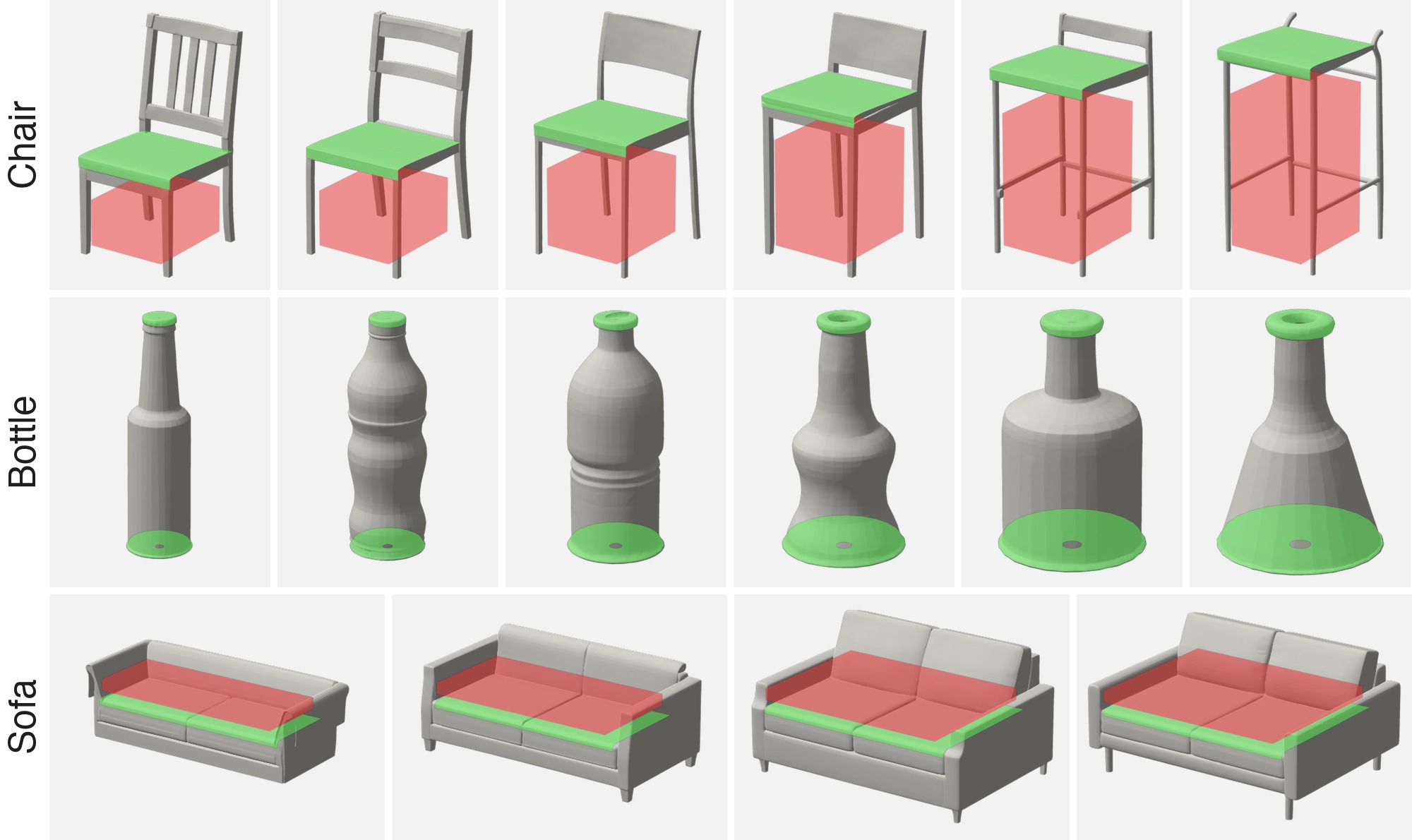}
    \caption{\textbf{Constraint sweeps.}
    The prompt stays fixed while the constraint changes over twelve steps; steps run from left to right (chair and bottle: 1, 3, 5, 7, 9, 12; sofa: 1, 5, 9, 12).
    The constraint is drawn translucently over each generated shape (green: hull, red: avoidance).
    Chair: the seat hull rises while the free space below it grows, and the dining chair becomes a bar stool.
    Bottle: the hull rings at the opening and the base widen, and the body follows.
    Sofa: the seat hull and the free space above it scale up together.
    Videos of all sweeps are in the supplement.}
    \label{fig:constraint_sweep}
\end{figure}

\subsection{Controlled generation}
Fig.~\ref{fig:comparison} compares the methods on eight hand authored constraints, with the full prompts listed in the supplement (Sec.~\extref{app:fig3_prompts}).
Each constraint fixes only some facts about the object, such as the frame of a rocking chair and the floor its rockers touch.
TRELLIS without geometry generates the right kind of object but places and sizes it on its own, SpaceControl often returns the constraint itself, a plain box for the sink or a chair without rockers, and Spice-E produces broken or different objects.
Arbor generates the object that the prompt names inside the given envelope, such as a rocking chair on the floor plane or a low car under the empty slab.
Fig.~\ref{fig:recon_comparison} asks whether stronger visual evidence is enough.
SPAR3D and Hunyuan3D-Omni receive a render of the source object and points sampled from the hull, and stay close to this image across seeds.
Arbor changes the object from seed to seed, for example the head of the hammer or the back of the chair, while it fills the hull and keeps out of the avoidance region.

Tab.~\ref{tab:metrics} shows the same pattern over the full splits.
Arbor has the highest Ctrl on both splits and respects nearly every avoidance region.
TRELLIS keeps the prompt but reaches little of the hull.
The constraint tokens move hull overlap close to that of the decoded source object, and prompt retrieval stays where it was.
SpaceControl reaches much of the hull but loses the prompt and the volume, because it tends to return the constraint itself.
Gradient pulls mass into the hull but often enters avoidance regions and leaves torn surfaces.
Against Gradient, the strongest baseline, the paired gain in Ctrl is about $0.2$ on both splits, with bootstrap intervals above zero.
Arbor also leads on the hull column alone, the one input that every method receives.
Given only this hull, like the baselines, Arbor still has the highest Ctrl on both splits.

The decoder adapters hardly change these scores, since they act below the $64^3$ grid of the control terms (Sec.~\ref{sec:experiments:detail}).

Arbor's meshes are as complex as those of TRELLIS, measured by the number of connected parts that each cover at least one percent of the surface.
In a user study with pairwise choices between rendered outputs, participants preferred Arbor over the baselines most often (Appendix~\extref{app:metrics}).

\subsection{Detail at the decoded surface}\label{sec:experiments:detail}
Tab.~\ref{tab:metrics} reads the structure at $64^3$, so it cannot tell whether a thin hull part survives decoding.
In Tab.~\ref{tab:detail256} we decode the same structures with the TRELLIS decoder and with the hull decoder and measure again at $256^3$.
The hull decoder recovers hull detail on almost every object and cuts the mean hull deviation by about a quarter; the gain vanishes at $64^3$, which explains the tie in Tab.~\ref{tab:metrics}.
With it, generated structures reach about the hull detail that the TRELLIS decoder recovers from ground truth latents, and Fig.~\ref{fig:decoder_detail} (left) shows the difference on one object.

\begin{table}[t]
    \centering
    \scriptsize
    \renewcommand{\arraystretch}{1.0}
    \setlength{\tabcolsep}{4pt}
    \newcommand{\best}[1]{\cellcolor{cvprblue!14}\textbf{#1}}
    \newcommand{\gt}[1]{\textcolor{black!50}{#1}}
    \newcommand{\sw}[1]{\textcolor{#1}{\rule{0.55em}{0.55em}}\,}
    \begin{tabular*}{\columnwidth}{@{\extracolsep{\fill}}lcccc@{}}
        \toprule
        & \multicolumn{2}{c}{\sw{hull}Hull at $256^3$ $\uparrow$} & \multicolumn{2}{c}{\sw{hull}Hull deviation $\downarrow$} \\
        \cmidrule(lr){2-3} \cmidrule(l){4-5}
        Method & Manual & Auto & Manual & Auto \\
        \midrule
        TRELLIS                 & 0.108 & 0.098 & 10.49 & 13.86 \\
        Arbor, TRELLIS decoder  & 0.554 & 0.662 & 1.28 & 0.91 \\
        \textbf{Arbor, hull decoder} & \best{0.699} & \best{0.774} & \best{0.94} & \best{0.69} \\
        \midrule
        \gt{GT latent, TRELLIS decoder} & \gt{0.714} & \gt{0.774} & \gt{0.95} & \gt{0.81} \\
        \gt{GT latent, hull decoder}    & \gt{0.779} & \gt{0.816} & \gt{0.81} & \gt{0.70} \\
        \bottomrule
    \end{tabular*}
    \caption{\textbf{Hull detail at $256^3$.}
    Hull overlap at the resolution of the decoded surface, and hull deviation, the mean distance from the hull surface to the generated surface in $256^3$ voxels.
    The two Arbor rows decode the same latents.}
    \label{tab:detail256}
\end{table}

\begin{figure}[t]
    \centering
    \includegraphics[width=\columnwidth]{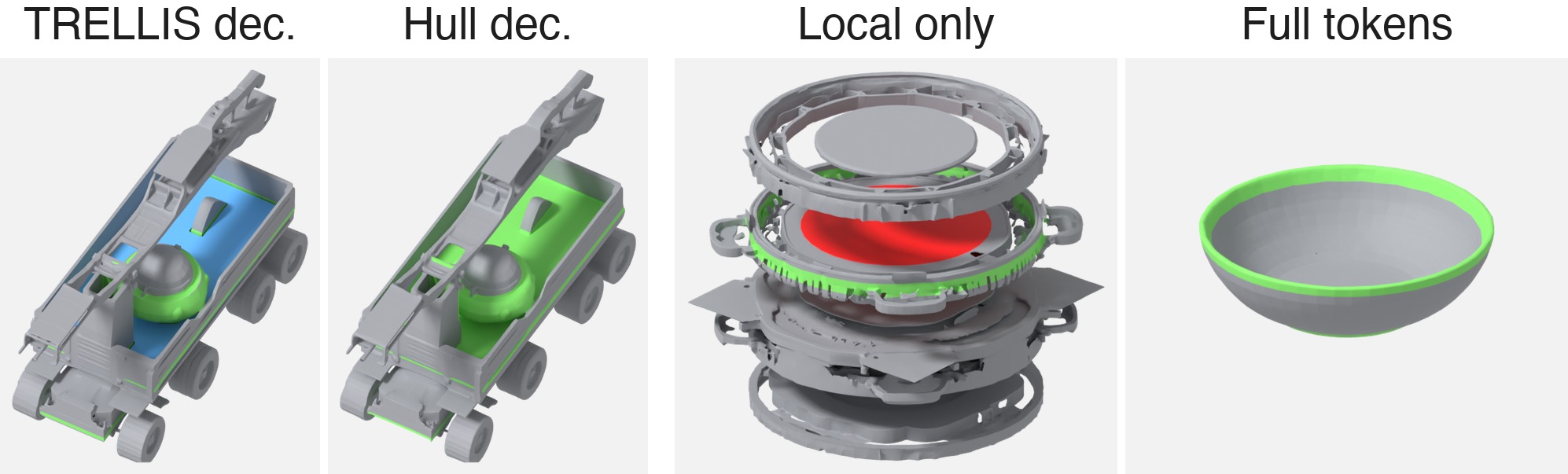}
    \caption{\textbf{Decoder and conditioning tokens.}
    Left: the same latent decoded twice. The hull is a thin floor plate and the walls of the truck bed; the TRELLIS decoder leaves most of it empty (\textcolor[rgb]{0.35,0.7,1.0}{blue}), while the hull decoder recovers it (green). Hull overlap at $256^3$ rises from 0.24 to 0.80 on this object.
    Right: the hull is a rim ring and a base disc, and the free space inside the bowl is an avoidance region. With only the local constraint tokens, Arbor-XL stacks rings and discs at and around the hull, joins them with no wall, and fills the avoidance region (red); with the full token set, the same constraint yields one coherent bowl.}
    \label{fig:decoder_detail}
    \label{fig:probe_tradeoff}
\end{figure}

\subsection{Analysis}\label{sec:experiments:ablations}

\textbf{Conditioning probes.}
Removing parts of the constraint tokens at test time (Tab.~\ref{tab:probe_findings}) shows what the trained model relies on.
Without positions the hull overlap falls far below that of the generator without geometry, and without roles it falls to its level.
Local tokens alone reach less of the hull and build much less of the object than the full set (Fig.~\ref{fig:probe_tradeoff}, right), so the global summary is needed next to them.

\begin{table}[t]
    \centering
    \scriptsize
    \renewcommand{\arraystretch}{1.0}
    \setlength{\tabcolsep}{0pt}
    \newcommand{\best}[1]{\cellcolor{cvprblue!14}\textbf{#1}}
    \newcommand{\sw}[1]{\textcolor{#1}{\rule{0.55em}{0.55em}}\,}
    \begin{tabular*}{\columnwidth}{@{\extracolsep{\fill}}lcccccc@{}}
        \toprule
        & \textbf{Full} & \makecell{Local\\only} & \makecell{No\\shape} & \makecell{No\\role} & \makecell{No\\pos.\ enc.} & \makecell{No\\geometry} \\
        \midrule
        \sw{hull}Hull$\uparrow$          & \textbf{.896} & .715 & .715 & .274 & .051 & .246 \\
        \sw{avoidance}Avoid$\uparrow$ & \textbf{.981} & .813 & .918 & .819 & .562 & .777 \\
        Vol.$\uparrow$ & \textbf{.717} & .254 & .666 & .678 & .293 & .586 \\
        \bottomrule
    \end{tabular*}
    \caption{\textbf{Conditioning probes.}
    What the trained model relies on: parts of the constraint tokens are removed at test time, on the 32 hand authored constraint sets with two seeds. Without positions, hull overlap falls far below no geometry (TRELLIS), and without roles to its level.}
    \label{tab:probe_findings}
\end{table}

\textbf{Sweeps.}
Fig.~\ref{fig:constraint_sweep} tests control on constraints outside the benchmark.
The prompt stays fixed while the constraint changes in small steps. A seat hull rises over a growing free region, the rings of a bottle widen, and a sofa seat grows together with the free space above it.
At every step Arbor follows the constraint and still generates a good object, as when the dining chair turns into a bar stool.
The teaser and the sweeps use hand authored constraints that are edited well beyond the geometry of their source objects.

\inputlocal{after_experiments.tex}
\section{Conclusion}\label{sec:conclusion}

Arbor lets artists state spatial requirements before generation, as constraint meshes that a frozen generator reads through a routed attachment.
From one set of constraints an artist can then generate a whole library of assets that all work with the same animation cycle in a game.
Among the methods we compare, it is the only one that combines a text prompt with coarse shapes, partial geometry and space that has to stay empty (Tab.~\ref{tab:capabilities}).
On Toys4K, it follows the constraints more closely than all compared methods and keeps the prompt and the complexity of the unconstrained generator.

Two limits remain.
A hull carries geometry but no meaning, so what the object builds inside it is left to the hull and the prompt.
Attaching part labels to the regions is the natural next step.
The resolution is set by the frozen generator, not by Arbor, and we look forward to applying it to generators with a higher resolution.
Appendix~\extref{app:limitations} lists further limits.

\inputlocal{after_conclusion.tex}

\label{arbor:mainend}
\section*{Acknowledgements}
The authors thank Stability AI for hosting Jan-Niklas Dihlmann as an intern during this work.
This work was funded by the Deutsche Forschungsgemeinschaft (DFG, German Research Foundation) under Germany's Excellence
Strategy, EXC number 2064/1, project number 390727645.
This work was supported by the German Research Foundation (DFG), SFB 1233, Robust Vision: Inference Principles and Neural
Mechanisms, TP 02, project number 276693517, and by the T\"ubingen AI Center.
The authors thank the International Max Planck Research School for Intelligent Systems (IMPRS-IS) for supporting
Jan-Niklas Dihlmann.

{\small\bibliographystyle{ieeenat_fullname}\bibliography{references}}

\ifx\arborpart\arborpartmain\else
  \clearpage
  \appendix
  
  \inputlocal{after_appendix.tex}
\fi
\fi

\end{document}